\documentclass{article} %
\usepackage{iclr2027_conference,times}
\iclrfinalcopy 

\def\arxivversion{1}

\usepackage{amsmath,amsfonts,bm}

\def\eqref#1{equation~\ref{#1}}

\def\1{\bm{1}}

\DeclareMathAlphabet{\mathsfit}{\encodingdefault}{\sfdefault}{m}{sl}
\SetMathAlphabet{\mathsfit}{bold}{\encodingdefault}{\sfdefault}{bx}{n}

\usepackage{color}
\usepackage{epsfig}
\usepackage{graphicx}
\usepackage{xparse} 

\usepackage{adjustbox}
\usepackage{array}
\usepackage{booktabs}
\usepackage{colortbl}
\usepackage{wrapfig}
\usepackage{hhline}
\usepackage{multirow}
\usepackage{wrapfig}
\usepackage{floatflt}

\usepackage{bm}
\usepackage{nicefrac}
\usepackage{microtype}

\usepackage{changepage}
\usepackage{extramarks}
\usepackage{fancyhdr}
\usepackage{lastpage}
\usepackage{setspace}
\usepackage{soul}
\usepackage{xspace}
\definecolor{verylightgray}{rgb}{0.95,0.95,0.95}
\usepackage{amsfonts}
\usepackage{amsmath}
\usepackage{amssymb}
\usepackage{soul}
\usepackage{siunitx}
\usepackage{colortbl}
\usepackage{xcolor}
\usepackage{adjustbox}
\usepackage{array}
\usepackage{tabularx}
\usepackage{pifont}
\definecolor{cvprblue}{rgb}{0.21,0.49,0.74}
\usepackage[pagebackref,breaklinks,colorlinks,allcolors=cvprblue]{hyperref}
\usepackage{cleveref} 
\usepackage[normalem]{ulem}

\usepackage{enumerate}
\usepackage{enumitem}  %
\usepackage{makecell}

\usepackage{pifont} %
\usepackage{subcaption}

\usepackage{algorithm,algpseudocode}

\usepackage[symbol]{footmisc}

\usepackage[most]{tcolorbox}

\usepackage{url}
\usepackage{tcolorbox}    
\usepackage{graphicx}
\usepackage{algorithm}
\usepackage{algpseudocode}   
\usepackage{subcaption}  

\usepackage[most]{tcolorbox}
\newtcolorbox{keyinsight}{
    enhanced,
    colback=blue!4,
    colframe=blue!55!black,
    boxrule=0.6pt,
    arc=3pt,
    left=8pt, right=8pt, top=5pt, bottom=5pt,
    before skip=8pt, after skip=10pt,
    fontupper=\small,
    before upper={\textbf{\color{blue!55!black}Key insight.}~}
}

\NewDocumentCommand{\heng}
{ mO{} }{\textcolor{red}{\textsuperscript{\textit{Heng}}\textsf{\textbf{\small[#1]}}}}

\NewDocumentCommand{\ember}
{ mO{} }{\textcolor{orange}{\textsuperscript{\textit{Ember}}\textsf{\textbf{\small[#1]}}}}

\newcommand{\inc}[1]{$_{\uparrow#1}$}   

\title{Beyond Entropy: Self-Diagnostic Multi-Role Token Optimization for Video Reasoning}

\author{%
  Yudong Han\textsuperscript{1,2,*},
  Yong Wang\textsuperscript{2,*},
  Zaiquan Yang\textsuperscript{3}, \\
  \textbf{Liang Lin\textsuperscript{4}},
  \textbf{Chongyang Tao},
  \textbf{Xiangxiang Chu\textsuperscript{2}},
  \textbf{Liyuan Pan\textsuperscript{1,\dag}}, \\[0.5em]
  \textsuperscript{1}Beijing Institute of Technology \quad
  \textsuperscript{2}DreamX Team \quad
  \textsuperscript{3}City University of Hong Kong \\
  \textsuperscript{4}Institute of Information Engineering, Chinese Academy of Sciences (CAS) \\[0.3em]
}

{

\footnotetext[1]{\ Equal contribution. Work done when Yudong's internship at DreamX Team.}
\footnotetext[2]{\ Corresponding author.}
}

\usepackage{colortbl}
\definecolor{MyDarkBlue}{rgb}{0,0.08,0.8}
\definecolor{MyDarkGreen}{RGB}{45,155,45}
\definecolor{MyDarkRed}{rgb}{0.8,0.02,0.02}
\definecolor{MyDarkOrange}{RGB}{166, 82, 0}
\definecolor{MyOrange}{rgb}{1.0, 0.4, 0.2}
\definecolor{MyPurple}{RGB}{111,0,255}
\definecolor{MyRed}{rgb}{0.8,0.0,0.0}
\definecolor{MyGold}{rgb}{0.75,0.6,0.12}
\definecolor{MyDarkgray}{rgb}{0.66, 0.66, 0.66}

\definecolor{main_table_green1}{RGB}{230,245,230}
\definecolor{main_table_green2}{RGB}{200,235,200}
\definecolor{main_table_green3}{RGB}{170,225,170}
\definecolor{main_table_green4}{RGB}{115,205,115}

\newcommand{\ours}{\textsc{PAPO}\xspace}

\makeatletter
\newcommand{\thickhline}{%
	\noalign {\ifnum 0=`}\fi \hrule height 1pt
	\futurelet \reserved@a \@xhline
}
\makeatother

\begin{document}

\maketitle
\ifdefined\arxivversion
    \lhead{Preprint}
\fi

\begin{abstract}

Reinforcement learning with verifiable rewards has substantially advanced multimodal reasoning, yet it remains fundamentally limited by ambiguous token-level credit assignment. While high-entropy token heuristics encourage possibility exploration, naively extending them to video reasoning tends to induce lengthy reasoning, as the model becomes overly reliant on high-entropy visual activations. Alternative approaches that rely on counterfactual-based visual token localization for credit assignment also tend to over-prioritize visual exploration at the expense of decisive reasoning cues for answer derivation, thereby exacerbating the interference from spurious visual nuances. Moreover, these methods employ static counterfactual strategies that fail to co-evolve with the policy during training. In this paper, we introduce DyCPO, a co-evolutionary framework that jointly optimizes reliable token selection and adaptive counterfactual intervention. It constructs a multi-role dependence metric to balance visual exploration and answer-relevance mining in token-wise contrastive learning, while suppressing exploration-only filler tokens and spurious visual noise. Rather than relying on static counterfactual priors, DyCPO dynamically derives counterfactual signals from the model's own successful and failed rollouts, enabling self-diagnostic analysis and co-evolution of the optimization objective with the policy. Extensive experiments on complex video reasoning and general video understanding benchmarks demonstrate consistent performance improvements, establishing DyCPO as a robust token-level credit assignment paradigm for multimodal reinforcement learning.
\end{abstract}

\addtocontents{toc}{\protect\setcounter{tocdepth}{-1}}

\section{Introduction}
\label{sec:intro}

Reinforcement learning (RL) with verifiable rewards~\citep{dapo, grpo}, exemplified by Group Relative Policy Optimization~\citep{grpo} (GRPO), has emerged as a pivotal paradigm for enhancing reasoning in large language models. However, GRPO uniformly distributes trajectory-level rewards across all tokens, leaving the critical challenge of token-level credit assignment unresolved. While recent work~\citep{DBLP:journals/corr/abs-2505-22617, DBLP:conf/nips/WangYGZLLDCYZLY25} identifies high-entropy tokens (approximately 20\%) as key decision points for exploration in text-only reasoning, the generalizability of this heuristic to video reasoning remains uncertain. Although recent multimodal RL studies~\citep{DBLP:journals/corr/abs-2604-01840, rezaei2026cppo, lookback, DBLP:journals/corr/abs-2606-03937} emphasize visual perception for finer-grained credit assignment, blindly amplifying visual signals risks introducing noise rather than informative signals. Meanwhile, on-policy distillation (OPD) methods~\citep{video-opsd, yuan2026vision, va-opd} offer fine-grained supervision but are hindered by their reliance on privileged annotations and limited scalability.

\newlength{\figH}
\setlength{\figH}{3.1cm}

\begin{figure*}[t!]
    \centering
    \begin{minipage}[b]{0.22\linewidth}
        \centering
        \includegraphics[height=\figH, width=\linewidth, keepaspectratio]{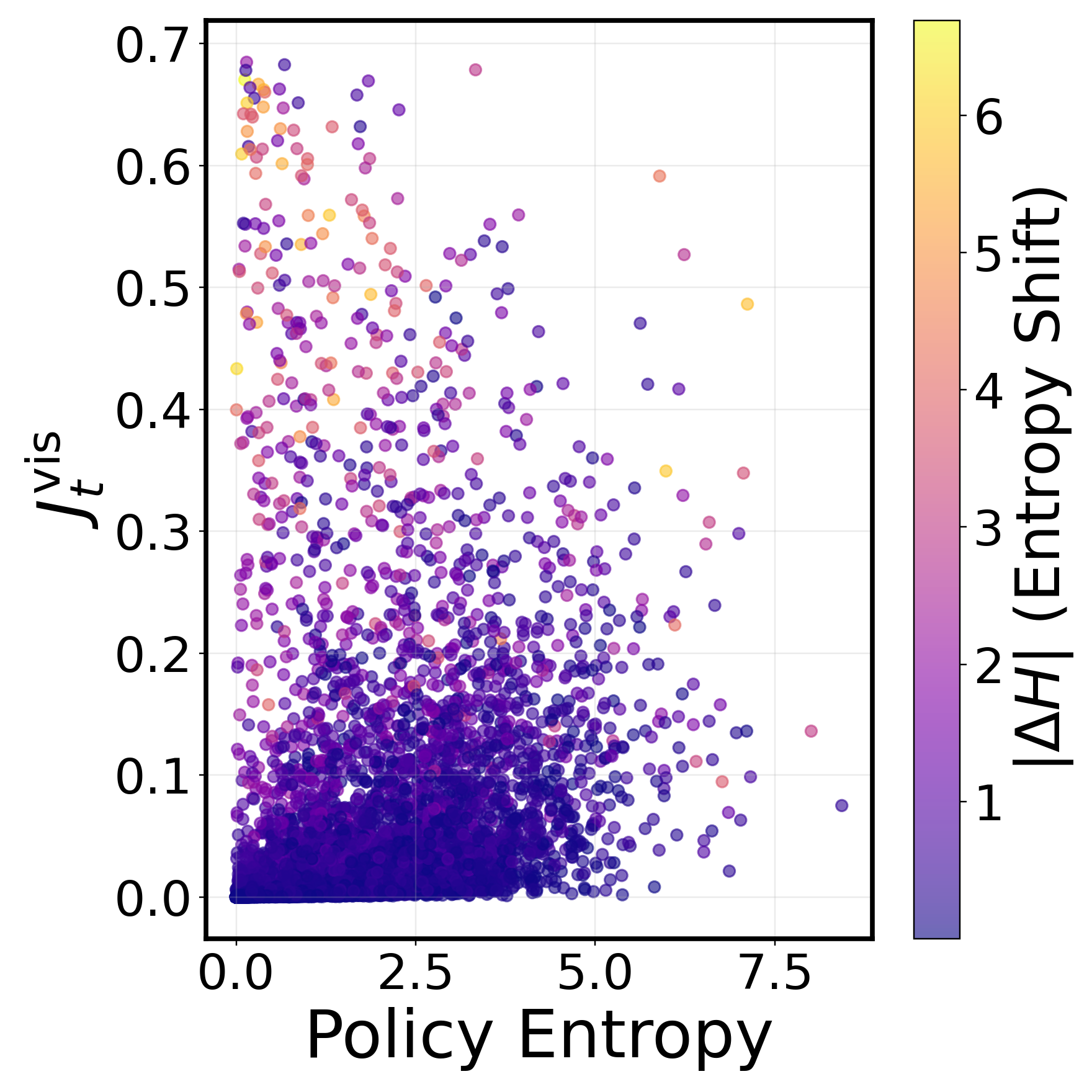}
    \end{minipage}%
    \hfill
    \begin{minipage}[b]{0.22\linewidth}
        \centering
        \includegraphics[height=\figH, width=\linewidth, keepaspectratio]{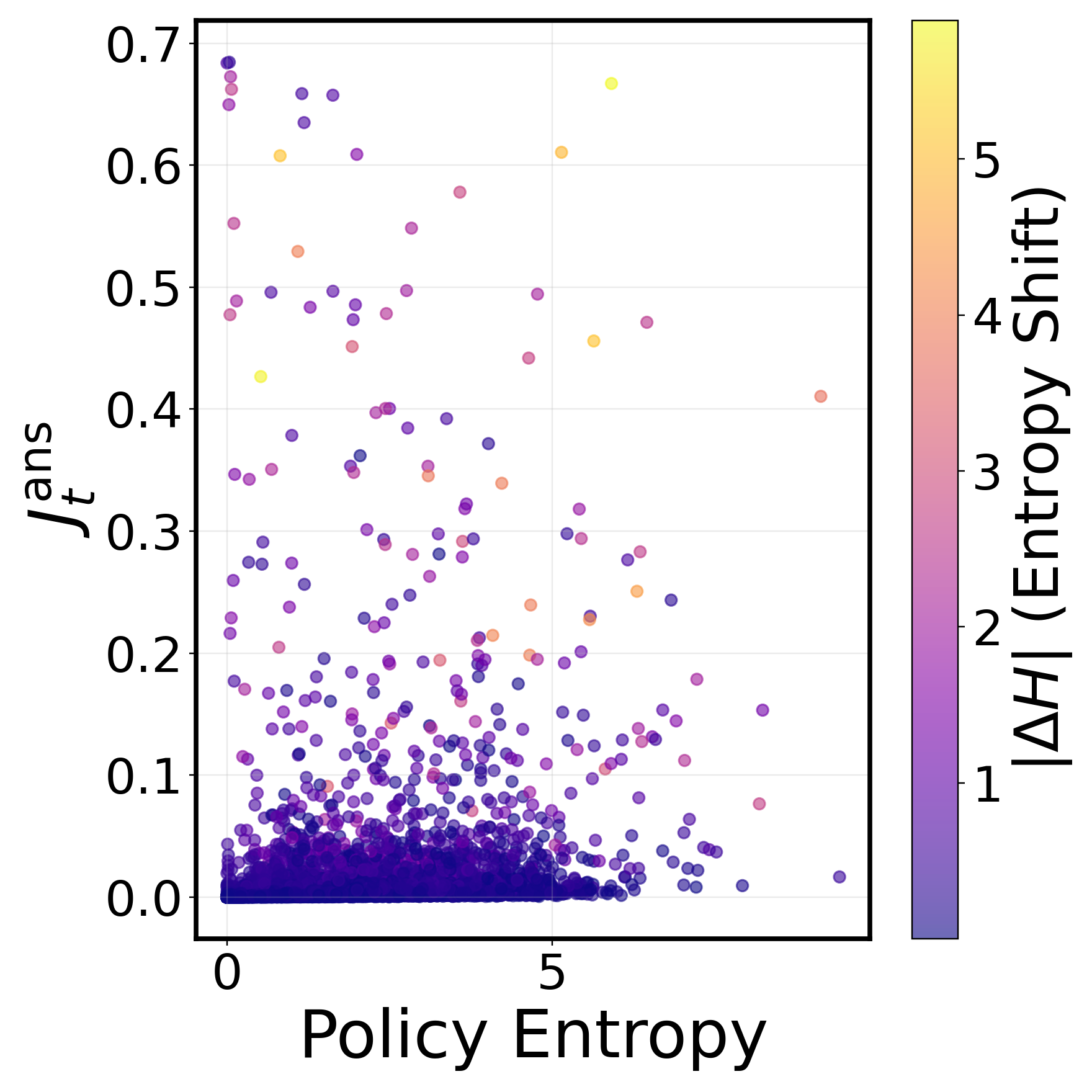}
    \end{minipage}%
    \hfill
    \begin{minipage}[b]{0.27\linewidth}
        \centering
        \includegraphics[height=\figH, width=\linewidth, keepaspectratio]{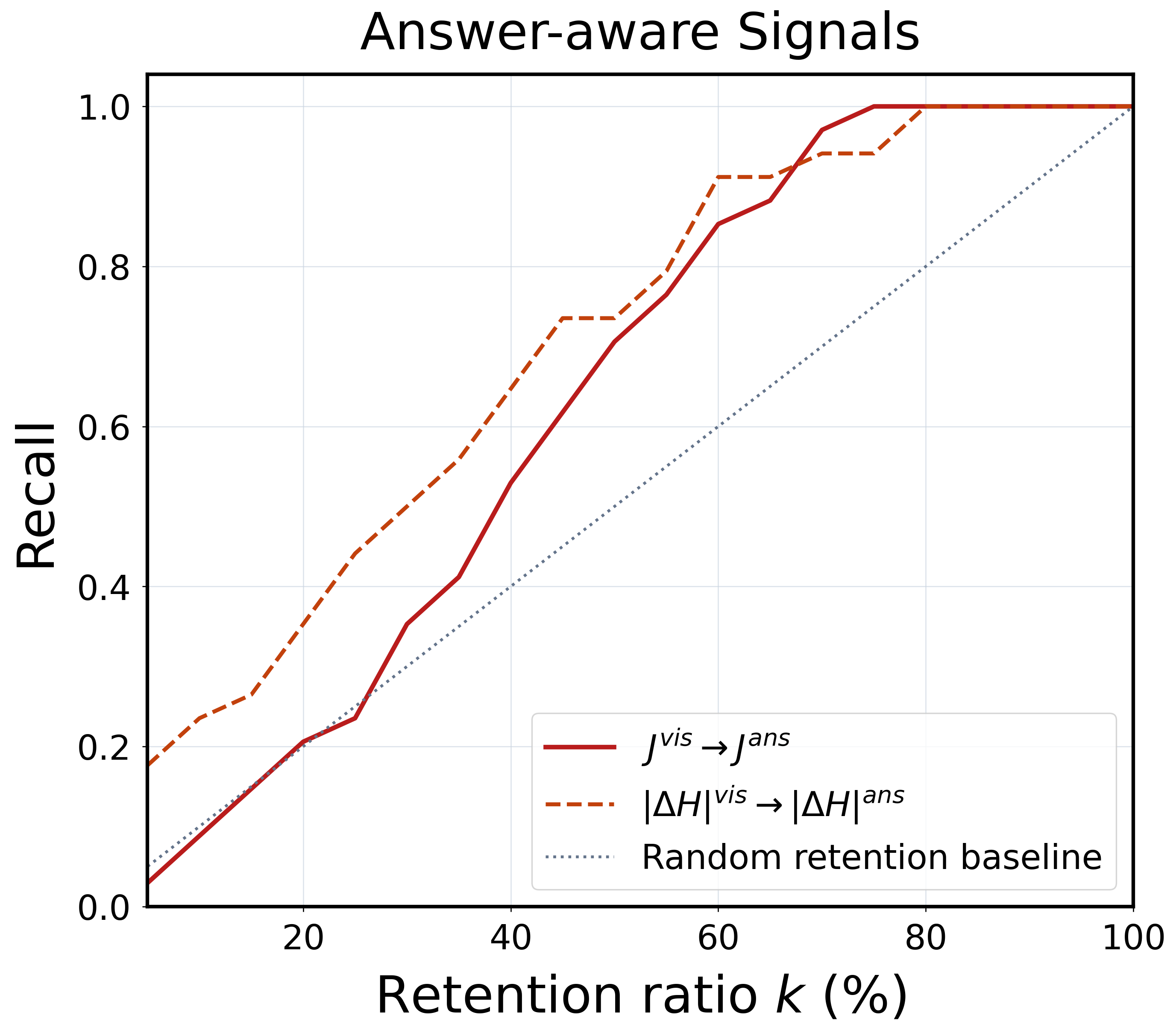}
    \end{minipage}%
    \hfill
    \begin{minipage}[b]{0.27\linewidth}
        \centering
        \includegraphics[height=\figH, width=\linewidth, keepaspectratio]{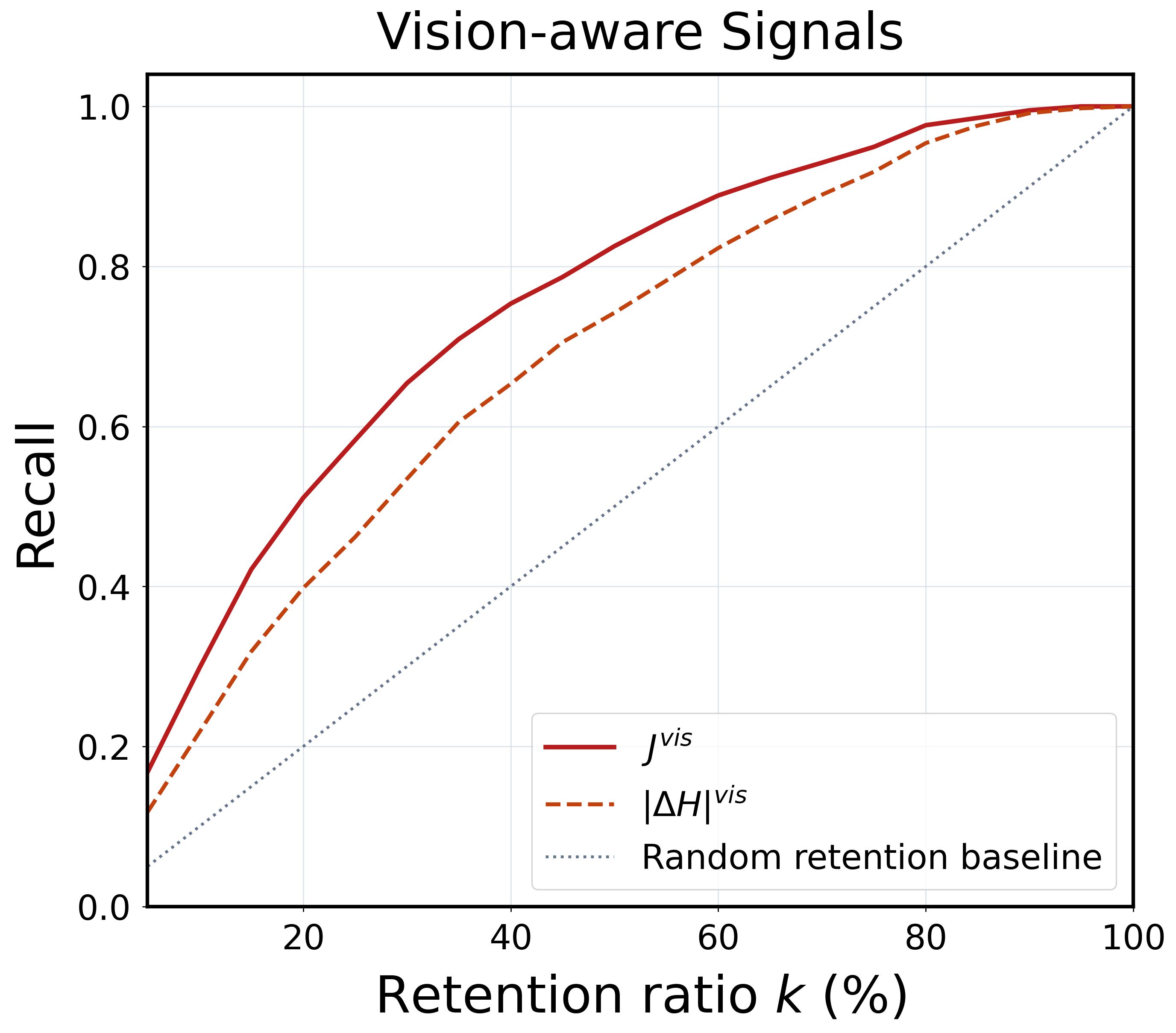}
    \end{minipage}

    \caption{\textbf{Left to right:} 
   (a, b) Numerous tokens with high vision-dependent and answer-dependent $\Delta H_{t}$ / $J_{t}$ reside in the low-entropy token set. (c, d) Token omission induced by single-selection metrics. Specifically, the recall in (c) is defined as $|A_{\text{ent}} \cap A_{\text{vis}}| / |A_{\text{vis}}|$, quantifying the coverage of visual-dependency tokens within the high-entropy selection. Similarly, the recall in (d) is calculated as $|A_{\text{vis}} \cap A_{\text{ans}}| / |A_{\text{ans}}|$, measuring the alignment between vision-sensitive and answer-critical tokens.}
    \label{fig:obs1}
\end{figure*}

\begin{figure*}[t]
  \centering
  \includegraphics[width=\textwidth]{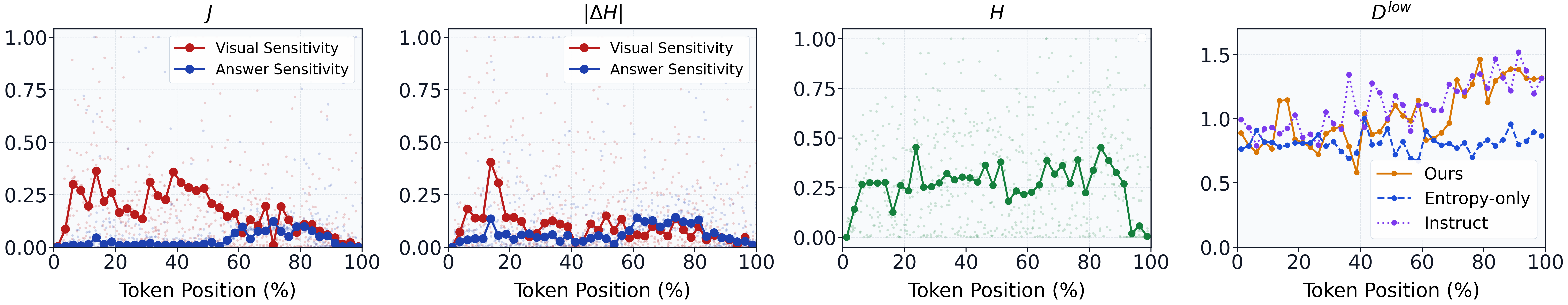}
  \vspace{-15pt}
  \caption{\textbf{Left to right:} (a, b) Vision- and answer-aware sensitivity across token positions, (c) entropy dynamics, and (d) reliance on towards low-entropy visual activations. Token position denotes the normalized decoding length.} 
  \label{fig:training_dynamics}
\end{figure*}

To investigate these issues, \textbf{(1)} we first demonstrate that naively rewarding token entropy would limit the video reasoning performance. Figure~2(c) illustrates that the instruction-tuned model without RL training exhibits a rise-then-fall entropy trend during decoding, while Figure~2(d) further reveals a progressively increasing reliance on low-entropy visual activations. Together, these dynamics reflect two distinct functional stages: during the early perception stage, the entropy of decoding token peaks as the model explore more fine-grained visual details. Consequently, in the later reasoning stage, once most visual facts are incorporated into the CoT prefix, the model shifts its focus to low-entropy activations, resulting in a drop in overall token entropy during the decision-making process. Rather than being redundant, these activations act as stable structural anchors that enforce spatio-temporal consistency and object representation, ultimately mitigating over-reasoning and hallucination. This finding extends the observations in~\citep{DBLP:journals/corr/abs-2511-19972}, providing further empirical substantiation for this mechanism. In contrast, the model optimized with the entropy-only (i.e., high entropy) setting consistently maintains a relatively low reliance on low-entropy visual activations throughout the decoding process, as indicated by the blue curve in Figure~2(d). Without converged visual grounding, this setting misdirects gradient signals toward exploratory tokens at the expense of decisive ones, resulting in noisy updates and the neglect of decisive content. These observations motivate us to move beyond naive entropy heuristics.

\textbf{(2)} As Figures~\ref{fig:obs1}(a) and (b) illustrate, relying on a single heuristic for token selection inevitably introduces critical blind spots. On the one hand, exclusively selecting the top-20\% highest-entropy tokens causes the model to miss nearly half of the tokens that are genuinely grounded in the video (as indicated by $J_{vis}$ and $|\Delta H|_{vis}$ in the right plot). On the other hand, focusing solely on vision-dependent tokens fails to capture the actual trajectory of the reasoning process. As the left plot ($J_{vis} \rightarrow J_{ans}$) demonstrates, selecting the top-20\% most vision-sensitive tokens captures merely 20\% of the signals required for the final answer, yielding performance no better than random selection. This observation demonstrates that, beyond mere entropy heuristics, ``seeing'' (perceptual grounding) and ``thinking'' (logical deduction) exhibit distinct token preferences. Consequently, this motivates us to evaluate tokens using decoupled \textit{vision-dependent} and \textit{answer-dependent} metrics.

\textbf{(3)} To further probe the divergence in token selection, we apply visual and answer counterfactual interventions to measure token sensitivity to input perturbations. As shown in Figure~\ref{fig:obs1}(a)(b), tokens with high $\Delta H_{t}$ or $J_{t}$ under visual counterfactuals concentrate in early decoding steps (0--40), with magnitudes monotonically decaying across the sequence. In contrast, answer-sensitive tokens localize to middle and late reasoning stages (60--80). The token cloud distributions in Figure~\ref{fig:combined_three} further corroborate their semantic complementarity: the former anchor on visual entities or attributes, while the latter mark logical pivots and conclusions, filtering out vision-irrelevant noise. These results collectively demonstrate that any single-role criterion is inherently insufficient for token-level credit assignment in video reasoning.

We therefore advocate for \emph{multi-role tokens} as the most informative and least noise-prone candidates for credit assignment: they are sensitive to both visual and answer counterfactuals while retaining high entropy. In this paper, we design a multi-role dependency score to quantify token importance and steer policy-gradient updates toward more robust optimization directions. A further critical observation is that random or static counterfactual strategies can induce model collapse when applied to dynamic video reasoning scenarios. We conjecture that fixed perturbation strategies are suboptimal because they fail to adapt to the evolving model state during training. To address this, we propose a state-conditioned mechanism that dynamically adjusts counterfactual generation. Specifically, we extract informative signals from rollout trajectories to guide this adaptive process by thoroughly mining fine-grained cues. These rollouts are generated by a proposal network that is updated in lockstep with the policy model at each training step.

Extensive experiments on knowledge reasoning, general understanding, spatial perception, long-video benchmarks, and video hallucination demonstrate that DyCPO achieves state-of-the-art or highly competitive performance across diverse tasks. Notably, on Video-Holmes, DyCPO attains an accuracy of 44.8\%, outperforming GPT-4o (42.0\%) by a significant margin. Comprehensive analyses and ablations confirm that jointly optimizing multi-role credit assignment and state-conditioned counterfactuals provides an effective, interpretable, and robust RL foundation for multimodal video reasoning.

\section{Related Work}
While reinforcement learning (RL) algorithms such as GRPO~\citep{DeepseekR1} or minimalist policy gradient baselines like GPG \citep{chu2026gpg} have have revolutionized textual reasoning or image-based reasoning~\cite{}, extending them to multimodal large language models (MLLMs) often neglects temporal dynamics. To bridge this gap, recent video-centric RL methods~\citep{video-r1, video-rts} incorporate temporally perturbed contrastive learning, introduce tool-based agent pipelines~\citep{video-o3} that decouple grounding from reasoning for long-form video understanding, or design scalable reward functions such as difficulty-aware regression~\citep{deepvideo-r1}. Nevertheless, these approaches predominantly rely on coarse sequence-level rewards or simple heuristics, failing to model fine-grained and reliable reasoning processes. Consequently, recent efforts have shifted toward improving token-level credit assignment via attribution modeling~\citep{video-ktr} or leveraging the teacher's privileged information to provide fine-grained supervision within on-policy distillation (OPD) frameworks~\citep{video-opsd}.

\section{Methods}

\subsection{Preliminaries}
\noindent\textbf{RL for video reasoning.}
Given a multimodal query $x = (\mathcal{V}, q)$ with video $\mathcal{V}$ and question $q$, the policy $\pi_\theta$ generates a reasoning trajectory $\mathbf{o} = (o_1, \dots, o_{M})$. Group Relative Policy Optimization (GRPO) samples $G$ candidate responses $\{\mathbf{o}_i\}_{i=1}^G \sim \pi_{\theta_{\mathrm{old}}}(\cdot \mid x)$ and computes group-normalized relative advantages $A_i = (R(\mathbf{o}_i) - \mu_R) / (\sigma_R + \delta)$, where $R(\mathbf{o}_i)$ is a scalar verifier reward, $\mu_R$ and $\sigma_R$ are the reward mean and standard deviation over the group, and $\delta > 0$ prevents numerical division by zero. The policy is updated by maximizing:
\begin{equation}
\mathcal{L}_{\mathrm{GRPO}}(\theta) = \mathbb{E}\left[\frac{1}{G}\sum_{i=1}^G \frac{1}{|\mathbf{o}_i|}\sum_{m=1}^{|\mathbf{o}_i|} \min\big(r_{i,m}A_i, \mathrm{clip}(r_{i,m}, 1{-}\epsilon, 1{+}\epsilon)A_i\big) - \beta D_{\mathrm{KL}}(\pi_\theta \parallel \pi_{\mathrm{ref}})\right],
\label{eq:grpo_loss}
\end{equation}
where $r_{i,m} = \pi_\theta(o_{i,m}\mid x,\mathbf{o}_{i,<m}) / \pi_{\theta_{\mathrm{old}}}(o_{i,m}\mid x,\mathbf{o}_{i,<m})$ is the importance sampling ratio, $\epsilon$ is the clipping threshold, and $\beta \ge 0$ penalizes divergence from the frozen reference policy $\pi_{\mathrm{ref}}$. While GRPO steers exploration via trajectory-level advantages $A_i$, its uniform weighting across decoding steps obscures fine-grained credit assignment, motivating our token-level analysis.

To diagnose token behavior, we introduce the following definitions.

\noindent\textbf{Token entropy.} We define the Shannon entropy of a distribution $P$ as $H(P) = -\sum_{w} P(w) \log P(w)$, and the Jensen--Shannon Divergence between distributions $P$ and $Q$ as:
\begin{equation}
J(P, Q) = \tfrac{1}{2}D_{\mathrm{KL}}(P \parallel M) + \tfrac{1}{2}D_{\mathrm{KL}}(Q \parallel M), \quad \text{where } M = \tfrac{P + Q}{2}.
\label{eq:jsd_general}
\end{equation}

\noindent\textbf{Reliance on low-entropy visual activations.} To quantify token grounding in stable visual cues, we define a reliance score on low-entropy visual tokens. We project layer-$l$ visual states $\mathbf{h}^{(l)}_v \in \mathbb{R}^d$ into the vocabulary space via the unembedding matrix $\mathbf{W}_U \in \mathbb{R}^{|\mathcal{Y}| \times d}$, obtaining $P^{(l)}_v = \mathrm{softmax}(\mathbf{W}_U \mathbf{h}^{(l)}_v)$. The deterministic visual anchor set $\mathcal{V}^{l}_{\mathrm{low}}$ and the visual reliance score $D^{\mathrm{low}}_{t}$ of token $o_t$ are defined as:
\begin{equation}
D^{\mathrm{low}}_{m} = \frac{1}{|\mathcal{L}|}\sum_{l\in\mathcal{L}}\sum_{v\in\mathcal{V}^{l}_{\mathrm{low}}}a^{(l)}_{m,v},\quad \mathcal{V}^{l}_{\mathrm{low}} = \left\{ v \in \mathcal{V} \mid H(P^{(l)}_v) \le \tau_{\mathrm{low}}^{(l)} \right\}
\label{eq:dep-low-entropy}
\end{equation}
where $\tau_{\mathrm{low}}^{(l)}$ is the bottom $q_{\mathrm{low}}$-th quantile threshold of visual entropies at layer $l$, and $a^{(l)}_{m,v} \in [0, 1]$ is the normalized cross-attention weight from token $o_m$ to visual position $v$ at layer $l$.

\noindent\textbf{Vision- and answer-aware sensitivity.} We probe token dependencies on visual and answer cues via counterfactual perturbations. For visual sensitivity, replacing video $\mathcal{V}$ with a perturbed counterfactual stream $\mathcal{V}'$ yields distribution $Q_m = \pi_\theta(\cdot \mid \mathcal{V}', q, \mathbf{o}_{<m})$, with shifts:
\begin{equation}
J^{\mathrm{vis}}_m = J(P_m, Q_m), \quad \Delta H^{\mathrm{vis}}_m = H(Q_m) - H(P_m).
\label{eq:vis_sens}
\end{equation}
For answer sensitivity, fixing $\mathcal{V}$ and contrasting verified answer $a^+$ against counterfactual alternative $a^- \in \mathcal{A} \setminus \{a^+\}$ yields conditional distributions $P^{\pm}_m = \pi_\theta(\cdot \mid \mathcal{V}, q, a^\pm, \mathbf{o}_{<m})$, inducing shifts:
\begin{equation}
J^{\mathrm{ans}}_m = J(P^+_m, P^-_m), \quad \Delta H^{\mathrm{ans}}_m = H(P^-_m) - H(P^+_m).
\label{eq:ans_sens}
\end{equation}
These metrics form the foundation for diagnosing token behavior, and are leveraged to optimize multi-role credit assignment in Section~\ref{sec:method}.

\subsection{Multi-Role Contrastive Credit Assignment}
\label{sec:method}

\noindent \textbf{Multi-role token selection.} As motivated above, we select \emph{multi-role tokens} as the optimal units for policy optimization, as they jointly capture sensitivity to visual grounding and answer derivation while filtering out noise-prone tokens with low informativeness. Building on the sensitivity metrics defined earlier, we construct a composite score to identify these tokens. Given a rollout $\mathbf{o}_i = \{ o_{i,1}, \dots, o_{i,M}\}$ generated by policy $\pi_\theta$, we first normalize the two sensitivity signals per rollout, then fuse them into a single joint score via a soft probabilistic OR:
\begin{equation}
c_{i,m}= \prod_{c \in \{\mathrm{vis}, \mathrm{ans}\}} \left[ \left(1 - \operatorname{Norm}\!\left(J^{c}_{i,m}\right)\right) \left(1 - \operatorname{Norm}\!\left(\lvert\Delta H^{c}_{i,m}\rvert\right)\right) \right]^{\alpha_{c}} \cdot H_{i,m},
\label{eq:joint-score}
\end{equation}
where $c \in \{ \mathrm{vis}, \mathrm{ans}\}$ indexes the visual and answer modalities, $J^{c}_{i,m}$ is the Jensen--Shannon divergence between the original and perturbed token distributions (i.e., $J^{\mathrm{vis}}_{i,m}$ for visual perturbation and $J^{\mathrm{ans}}_{i,m}$ for answer perturbation), $\Delta H^{c}_{i,m}$ denotes the entropy shift induced by the corresponding counterfactual intervention, $\operatorname{Norm}(\cdot)$ denotes min-max normalization within the rollout, $\alpha_{c} \in [0,1]$ with $\sum_{c} \alpha_{c}=1$ balances the contribution of visual exploration and reliable grounding, and $H_{i,m}$ denotes the entropy of the token distribution at step $m$, which reweights the fused sensitivity score by the token's informativeness.
Finally, the top-$k$ response tokens by $c_{i,m}$ form the multi-role set:
\begin{equation}
\mathcal{C} =
\left\{ m \,\middle|\, \frac{\operatorname{Rank}(c_{i,m})}{|\mathbf{o}_i|} \leq k \right\},
\label{eq:perception_topk}
\end{equation}
where $\operatorname{Rank}(\cdot)$ denotes the descending rank and $k \in (0, 1]$ the retention ratio. The resulting binary mask $M_{i,m} = \mathbb{I}\big[m\in\mathcal{C}\big]$ dynamically identifies the most informative and reliable multi-role tokens.

\noindent\textbf{Overall objective and policy update.}
Building on the multi-role token set $\mathcal{C}$ defined in Eqn.~\ref{eq:perception_topk}, we introduce a contrastive regularization term to reinforce informative and reliable token during policy update. Unlike prior work that directly reweights tokens~\citep{video-ktr}, our approach separates selected tokens from unselected ones by aligning their rollout prediction distributions with an information-preserving counterfactual view $\pi_\theta^{+}$ (with slight jitter $\mathcal{V}^+$) and distancing them from an information-removing view $\pi_\theta^{-}$ (with crucial visual evidence $\mathcal{V}^-$ removed). Aligning with $\pi_\theta^{+}$ stabilizes training and makes the model robust to small input changes, while distancing from $\pi_\theta^{-}$ prevents the model from relying on irrelevant or misleading features. For each selected token, the contrastive loss is defined as:
\begin{equation}
\mathcal{L}^{\mathrm{MRC}}_{i,m}
= M_{i, m} \cdot \operatorname{softplus}\!\left(
\frac{D_{\mathrm{KL}}\big(\pi_\theta(o_{i,m}) \,\|\, \pi_\theta^{+}(o_{i,m})\big)-D_{\mathrm{KL}}\big(\pi_\theta(o_{i,m}) \,\|\, \pi_\theta^{-}(o_{i,m})\big)}{\tau}
\right),
\label{eq:mrc-loss}
\end{equation}
where $\tau$ is a temperature parameter and $\operatorname{softplus}(x)=\log(1+e^x)$. The token-level loss $\mathcal{L}^{\mathrm{MRC}}_{i,m}$ smoothly vanishes once the model achieves the desired separation between the positive and negative counterfactual views on the selected multi-role tokens. To avoid reinforcing spurious patterns from failed rollouts, we gate the loss exclusively on successful samples with positive advantages ($A_i>0$) and directly incorporate it into the standard GRPO loss as the final training objective:
\begin{equation}
\mathcal{L}_{\mathrm{total}} = \mathcal{L}_{\mathrm{GRPO}}(\theta) + \frac{\lambda_{\mathrm{MRC}}}{|\mathcal{C}|} \sum_{m \in \mathcal{C}} \mathbb{I}(A_i > 0) \cdot \mathcal{L}^{\mathrm{MRC}}_{i,m},
\label{eq:loss_func}
\end{equation}
where $\lambda_{\mathrm{MRC}}$ balances the weight between outcome-driven policy optimization and token-wise self-distillation.

\subsection{Self-Diagnostic Evolving Optimization}
\begin{wrapfigure}{r}{0.36\textwidth} 
    \centering
    \includegraphics[width=\linewidth]{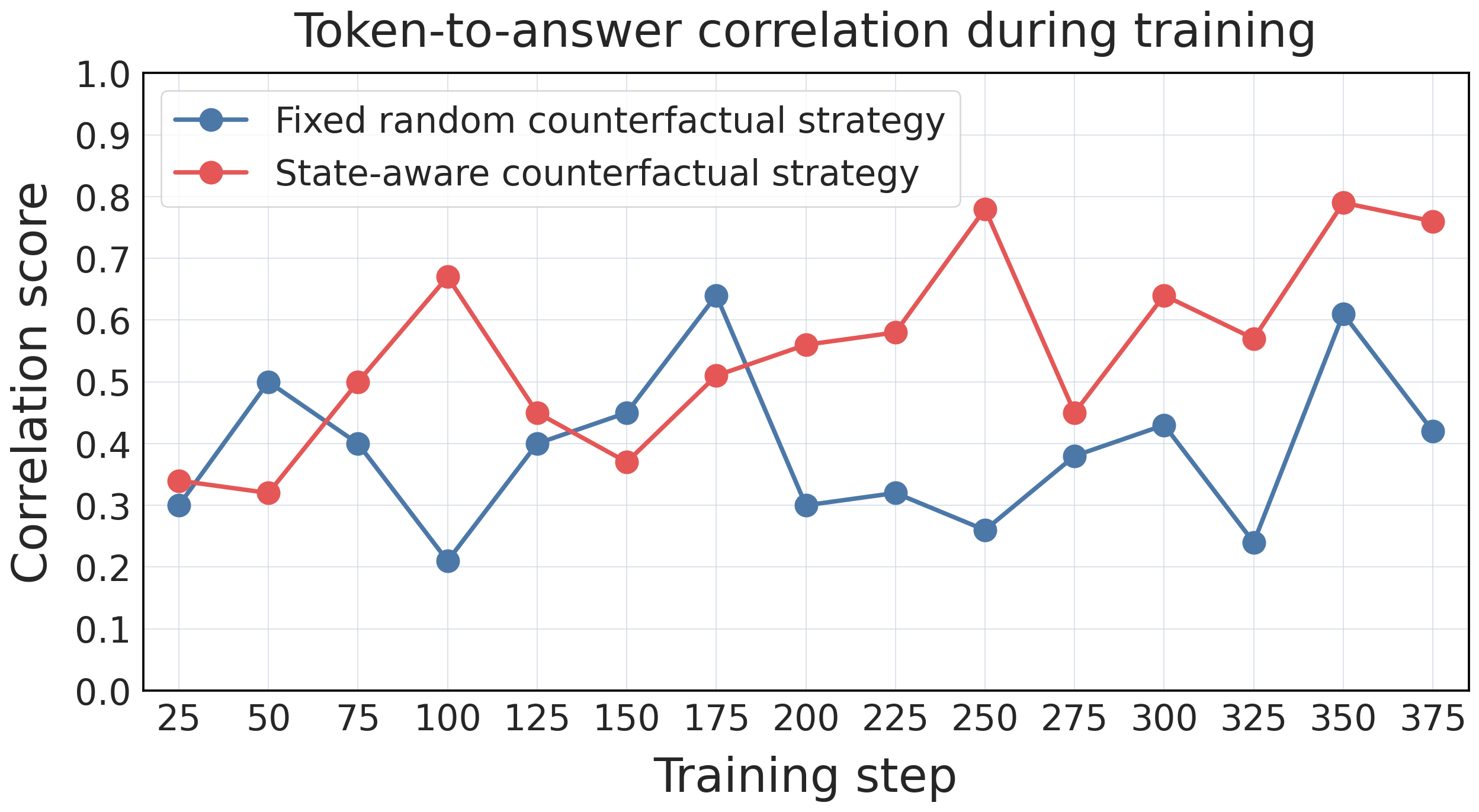}
    \vspace{-17pt}
    \caption{Training dynamics of the correlation between activated tokens and answer-relevant reasoning cues.}
    \label{fig:dynamics}
\end{wrapfigure}

Model optimization is inherently dynamic, and counterfactual interventions play a non-trivial role in identifying salient tokens. This motivates examining whether static counterfactual strategies can align with such dynamic training processes. As shown in Figure~\ref{fig:dynamics}, the fixed random masking baseline exhibits relatively low and volatile token-to-answer correlation, while our state-aware strategy achieves consistently higher correlation, particularly in later training stages, suggesting its capacity to co-evolve with the policy and prioritize answer-relevant visual clues. This indicates the value of adaptive counterfactual strategies for effective long-video reasoning.

\noindent \textbf{Rollout-contrastive modeling.} Specifcially, we mine informative signals from rollout trajectories, observing that visual attention divergence between correct and incorrect rollouts highlights reasoning-relevant cues, while answer segments contain the visual evidence the model converges on. Motivated by this, we model dynamics from two perspectives: (1) \textit{Dynamic visual counterfactuals}: We introduce a rollout-contrastive keyframe localization method, leveraging frame-level attention discrepancies between correct and incorrect rollouts as reasoning signals. Given $G$ rollouts from a proposal model $\pi_{\phi}$, we categorize them into $\mathcal{R}_r = \{\mathbf{o}_i \mid A(\mathbf{o}_i) \geq 0\}$ and $\mathcal{R}_w = \{\mathbf{o}_i \mid A(\mathbf{o}_i) < 0\}$ based on advantage values.

For each rollout trajectory $\mathbf{o}_i = \{o_{i,1}, \dots, o_{i,M}\}$, we track the attention weights from response tokens to visual frames. Motivated by prior findings~\citep{DBLP:conf/naacl/ZhangQSYYXWGTY25} demonstrating that attention converges to semantically relevant regions in intermediate layers but diverges in deeper ones, we compute the aggregated attention from the $m$-th token to the $t$-th frame $\mathcal{V}_t$ by averaging over a selected subset of model layers:
\begin{equation}
    \mathcal{A}^{(i)}_{m,t} = \frac{1}{H \cdot |\mathcal{V}_t|} \sum_{h=1}^{H} \sum_{v \in \mathcal{V}_t} a^{(h)}_{m,v},
    \label{eq:attention-weight}
\end{equation}
where $H$ is the number of attention heads, $|\mathcal{V}_t|$ denotes the number of visual tokens in frame $\mathcal{V}_t$, and $a^{(h)}_{m,v}$ is the raw attention weight from token $m$ to visual patch $v$ at head $h$. We first select the top-$K_1$ response tokens with the strongest visual grounding to frame $\mathcal{V}_t$, and average their attention scores to obtain the representative response-to-frame attention score:
\begin{equation}
    \bar{\mathcal{A}}^{(i)}_t = \frac{1}{K_1} \sum_{m \in \Phi^{(i)}_t} \mathcal{A}^{(i)}_{m,t}, \quad \text{where } \Phi^{(i)}_t = \operatorname{TopK}_{m \in \{1, \dots, M\}} \left( \mathcal{A}^{(i)}_{m,t},\; K_1 \right).
    \label{eq:hierarchical-token-attn}
\end{equation}

Subsequently, to pinpoint the video frames predominantly attended to by the entire response $\mathbf{o}_i$, we apply a second Top-$K_2$ selection across the video temporal dimension:
\begin{equation}
    \Psi^{(i)} = \operatorname{TopK}_{t \in \{1, \dots, T\}} \left( \bar{\mathcal{A}}^{(i)}_t,\; K_2 \right),
    \label{eq:topk-frames}
\end{equation}
where $\Psi^{(i)}$ represents the indices of the $K_2$ most visually salient frames for rollout $\mathbf{o}_i$. 

To construct a contrastive set of keyframes, we extract the union of identified focal frames within each respective group and define the critical region set $\mathcal{U}^\ast$ via set difference:
\begin{equation}
 \mathcal{U}^\ast = \Psi^+ \setminus \Psi^-, \quad \text{where } \Psi^+ = \bigcup_{\mathbf{o}_i \in \mathcal{R}_{r}} \Psi^{(i)}, \;\; \Psi^- = \bigcup_{\mathbf{o}_i \in \mathcal{R}_{w}} \Psi^{(i)}.
\label{eq:contrastive-set}
\end{equation}

This formulation interprets the gap between the visual focus of correct and incorrect rollouts as dense, region-level supervision derived from sparse outcome rewards. However, fixing a static keyframe set throughout training may compromise stability and generalization, as the policy's visual attention often shifts as its reasoning capabilities evolve. We therefore linearly anneal the keyframe count $K_2$ over the course of training, a curriculum that allows the model to first master fine-grained visual grounding before tracking broader temporal spans, smoothly extending contrastive self-distillation from static perception to temporal reasoning. (2) \textit{Answer contrastive counterfactuals}: We extract the correct and incorrect answer texts from the current rollouts as privileged information. Unlike directly sampling random answers from incorrect options, our strategy aligns with the model's own learning state, enabling it to better identify its current confusion or weaknesses, and thus further update gradients along this direction.

\noindent \textbf{EMA Self-Teacher.} 
We construct the proposal model $\pi_{\phi}$ via an Exponential Moving Average (EMA) of the parameters of policy model $\pi_{\theta}$, eliminating the need for a separately pretrained model. After each optimization step, the proposal parameters are updated according to:
\begin{equation}
    \pi_{\phi} \leftarrow (1 - \alpha) \pi_{\phi} + \alpha \pi_{\theta},
\end{equation}
where $\alpha \in (0, 1)$ is the EMA momentum coefficient. The proposal model $\pi_\phi$ is updated exclusively via exponential moving average (EMA) without backpropagation, serving as a slowly evolving rollout-contrastive anchor for self-distillation of the policy model. This setup prevents overly aggressive target drifts, enabling the policy model $\pi_\theta$ to fully adapt within a relatively stable counterfactual phase, thereby preserving rich temporal context. 

\begin{table*}[t!]
    \centering
    {\small
    \caption{\textbf{Performance Comparison.} \textbf{Bold} indicates the best results. $^\dagger$ marks results from original papers under different evaluation protocols. ``Data Size'' refers to the RL training data scale. The default video input consists of 32 frames.}
    \vspace{-5pt}
    \label{tab:sota_comparison}
    \resizebox{\textwidth}{!}{
    \setlength{\tabcolsep}{5pt}
    \renewcommand{\arraystretch}{1.2}
    \begin{tabular}{l|c|ccccc|cc}
         \toprule
         \rowcolor[gray]{.9}
        \textbf{Model} & \textbf{Data Size} & \textbf{VSI.} & \textbf{VideoMMMU} & \textbf{MMVU} & \textbf{Video-Holmes} & \textbf{MV.} & \textbf{TempC.} & \textbf{VideoMME} \\ 
         \midrule
        
        \multicolumn{9}{c}{\textit{Proprietary Models}}\\
        \midrule
        GPT-4o~\cite{gpt4o} & -- & 34.0 & 61.2 & 75.4 & - & - & - & 71.9 \\ 
        \midrule
        
        \multicolumn{9}{c}{\textit{Open-Source Models}}\\
        \midrule
        VideoLLaMA2~\cite{VideoLLaMA2} & -- & - & - & 44.8 & - & 54.6 & - & 47.9 \\
        LongVA-7B~\cite{LongVA} & -- & 29.2 & 23.9 & - & - & - & 56.9 & 52.6 \\
        VILA-1.5-8B~\cite{VILA} & -- & 28.9 & 20.8 & - & - & - & 58.8 & - \\
        LLaVA-OneVision-7B~\cite{LLaVAOneVision} & -- & 32.4 & 33.8 & 49.2 & - & 56.7 & - & 58.2 \\
        mPLUG-Owl3-8B~\cite{mPLUGOwl3} & -- & - & - & - & - & 54.5 & - & 53.5 \\
        Qwen2.5-VL-7B~\cite{Qwen25VL} & -- & 31.8 & 47.4 & 61.3 & - & 59.4 & 69.2 & 52.8 \\ 
        \midrule
        
        \multicolumn{9}{c}{\textit{Concurrent R1-based Models}}\\
        \midrule
        Video-R1~\textsubscript{\textcolor{gray}{NeurlPS{'}\,25}}~\cite{video-r1} & 260K & 35.8 & 52.3 & 63.8 & 36.5 & 63.9 & 72.6 & 59.3 \\
        VideoChat-R1~\textsubscript{\textcolor{gray}{NeurlPS{'}\,25}}~\cite{video-r1} & 18K & - & - & 64.2 & - & 63.1 & 72.9 & - \\
        TW-GRPO~\cite{VideoChatR1} & 1K & - & - & 65.8 & - & 63.3 & 73.3 & - \\
        Video-RTS~\textsubscript{\textcolor{gray}{ICLR{'}\,26}}~\cite{video-rts}    & 6K   & -    & 52.7 & 66.4 & 40.7 & - & - & 55.1 \\
        Video-RFT~\textsubscript{\textcolor{gray}{NeurlPS{'}\,25}}~\cite{video-rft} & 102K & 36.8 & 52.3 & 68.5 & - & 62.1 & 73.2 & 59.8 \\
        APPO~\textsubscript{\textcolor{gray}{CVPR{'}\,26}}~\cite{appo} & 260K & 32.7 & - & - & 40.8 & 64.6 &  71.9 & 59.8 \\
        Video-KTR~\textsubscript{\textcolor{gray}{ICLR{'}\,26}}~\cite{video-ktr} & 16.5K & 36.8 & 52.6 & 65.9 & 41.6 & - & 73.4 & 60.3 \\
        Ours (7B) & 3K & \cellcolor{blue!10}\textbf{39.2} & \cellcolor{blue!10}\textbf{53.6} & \cellcolor{blue!10}\textbf{68.5} & \cellcolor{blue!10}\textbf{44.8} & \cellcolor{blue!10}\textbf{65.8} & \cellcolor{blue!10}\textbf{74.2} & \cellcolor{blue!10}\textbf{62.6}\\ 
        \bottomrule
    \end{tabular}}
    }
\end{table*}
\section{Experiments}
\subsection{Experimental Setup}
\textbf{Training datasets and benchmarks.}
For Supervised Fine-Tuning (SFT), we leverage VideoRFT-CoT-102K to establish fundamental reasoning capabilities. For RL training, we curate two 3K subsets from the raw Video-R1 dataset~\citep{video-r1} via difficulty-based filtering to train the Qwen-2.5-VL and Qwen-3-VL models, respectively. This strategy retains only samples with verifiable answers while excluding trivially easy or excessively hard instances, ensuring that the RL data difficulty aligns with model capabilities. We evaluate our approach on seven challenging benchmarks: (1) \textbf{Video-Holmes}~\citep{video-holmes}, \textbf{VideoMMMU}~\citep{video-mmmu}, and \textbf{MMVU}~\citep{mmvu} for temporal, causal, and knowledge-intensive reasoning; (2) \textbf{TempCompass}~\citep{tempcompass} and \textbf{VideoMME}~\citep{videomme} for general capability; (3) \textbf{LVBench}~\citep{lvbench} for long-video comprehension; (4) \textbf{VSIBench}~\citep{vsibench} for spatial perception; and (5) video hallucination diagnosis \textbf{VidHalluc}~\citep{vidhalluc} .

\textbf{Implementation details.}
We use Qwen2.5-VL-7B-Instruct as the backbone for public benchmark evaluation and train all models on $16\times$ NVIDIA H20 GPUs (96GB each). For fair comparison, we follow the Video-R1 training protocol, and all ablations adopt identical configurations. During training, each video is uniformly sampled into 16 frames under a visual token budget of $128\times 28\times 28$ by default. We optimize the model with a learning rate of $1\mathrm{e}{-6}$ and a global batch size of 64. At inference, we increase the temporal input to 32 frames with resolution $256\times 28\times 28$. The RL pipeline is built on \texttt{verl}. More details are provided in the Appendix. 

\textbf{Baselines.}
We evaluate our method against a comprehensive set of state-of-the-art baselines with similar scale, including LLaVA-OneVision~\citep{LLaVAOneVision}, VILA-1.5~\citep{VILA}, Qwen2.5-VL~\citep{Qwen25VL}, Video-R1~\citep{video-r1}, Video-RTS~\citep{video-rts}, VideoRFT~\citep{video-rft}, TW-GRPO~\citep{tw-grpo}, Video-KTR~\citep{video-ktr}, APPO~\citep{appo} and on-policy distillation method Video-OPSD~\citep{video-opsd}.

\begin{table*}[t]
\centering
\small
\renewcommand{\arraystretch}{1.1}
\setlength{\tabcolsep}{6pt}
\caption{Ablation study of different scoring metrics across MMVU-MC, VSI, VideoMMMU, Video-Holmes, and VideoMME over three backbone architectures.}
\label{tab:score_ablation}
\resizebox{\textwidth}{!}{%
\begin{tabular}{clccccccccccc}
\toprule
\multirow{2}{*}{\textbf{Model}} & \multirow{2}{*}{\textbf{Method}} 
& \multirow{2}{*}{\textbf{MMVU}} 
& \multirow{2}{*}{\textbf{VSI}} 
& \multicolumn{4}{c}{\textbf{VideoMMMU}} 
& \multirow{2}{*}{\textbf{Video-Holmes}} 
& \multicolumn{4}{c}{\textbf{VideoMME}} \\
\cmidrule(lr){5-8}\cmidrule(lr){10-13}
& & & & Perc. & Compre. & Adapt. & Avg. & & Short & Medium & Long & Avg. \\
\midrule

\multirow{5}{*}{\raisebox{-0.5ex}{\rotatebox[origin=c]{90}{\begin{tabular}{c} Qwen2.5-VL \\ 7B \end{tabular}}}}
& GRPO                                                         & 65.3          & 35.9       & 61.7   & 47.3      & 41.3   & 50.1          & 40.9 & 70.0 & 58.4 & 49.5 & 59.3 \\
& wo  $\Delta H_{ans}$/ $J_{ans}$           & 64.3          & 36.5       & 61.0   & 47.7    & 41.7   & 50.1            & 41.2   & 71.3  & 59.7  & 50.4  & 60.5   \\
& wo  $\Delta H_{vis}$/ $J_{vis}$             & 67.2          & 37.9       & 62.3   & 47.0    & 37.0 & 48.8          & 42.1   & 70.7 &   57.8  & 49.5  & 59.3 \\
& wo $H_{t}$                                                 & 69.0          & 39.1       & 64.0   & 51.7   & 42.7   & 52.8            & 44.2   & 73.3 & 60.5 & 50.5 & 61.4 \\
& Our Full Model                                          & 68.5          & 39.2 & 63.7   &  52.3  & 43.3   & 53.1 & 44.8 & 74.3 & 61.5 & 52.0 & 62.6 \\
\midrule

\multirow{5}{*}{\raisebox{-0.5ex}{\rotatebox[origin=c]{90}{\begin{tabular}{c} Qwen3-VL \\ 2B \end{tabular}}}}
& GRPO                                                         & 60.1 & 47.4     & 60.0    & 42.3   & 29.0   & 43.8   & 35.5 & 69.7     & 52.8 & 46.2 & 56.2 \\
& wo  $\Delta H_{ans}$/ $J_{ans}$           & 60.2   & 48.1   & 58.7   & 46.3   & 35.0    & 46.7   & 36.1   & 68.4   & 51.9   & 47.2   & 55.8   \\
& wo  $\Delta H_{vis}$/ $J_{vis}$             & 59.5   & 46.8   & 57.3   & 47.3   & 34.3    & 46.3    & 37.4   & 68.2   & 51.3   & 46.5   & 55.3   \\
& wo $H_{t}$                                                 & 61.1   & 48.1   & 58.3   & 48.7   & 35.3    & 47.4   & 37.1   & 70.2   & 53.6    & 47.9   & 57.2 \\
& Our Full Model                                         & 61.6     & 49.4  & 60.0    & 49.3  & 36.3    & 48.5   & 37.7   & 70.5   & 54.3    & 48.4   & 57.7 \\
\midrule

\multirow{5}{*}{\raisebox{-0.5ex}{\rotatebox[origin=c]{90}{\begin{tabular}{c} Qwen3-VL \\ 4B \end{tabular}}}}
& GRPO                                                                    & 66.9 & 46.2 & 73.0       & 61.0 & 41.0 & 58.3 & 41.5        & 72.9  & 60.2 & 49.0 & 60.7 \\
& wo  $\Delta H_{ans}$/ $J_{ans}$                           & 66.2   & 47.8   & 73.3   & 61.3   & 42.7   & 59.1   & 39.1   & 72.4     & 60.9   & 51.2   & 61.5   \\
& wo  $\Delta H_{vis}$/ $J_{vis}$                              & 65.1   & 47.1   & 71.0   & 62.7   & 42.3   & 58.7   & 40.4   & 73.2    & 59.3   & 49.5   & 60.7   \\
& wo $H_{t}$                                                            & 65.6   & 47.8   & 74.3   & 62.3   & 44.7   & 60.4   & 41.2   & 74.5     & 60.2   & 49.9   & 61.5   \\
& Our Full Model                                                     & 67.0 & 48.3 & 74.7      & 63.0 & 45.3 & 61.0        & 42.2   & 74.4     & 61.9 & 53.0 & 63.1 \\
\bottomrule

\end{tabular}%
}
\vspace{0.5em}
\end{table*}

\subsection{Main Results}
\paragraph{Performance comparison with state-of-the-art models.}

As shown in Table~\ref{tab:sota_comparison}, our method consistently sets new state-of-the-art results across three reasoning and two general video understanding benchmarks. On \textbf{Video-Holmes}, which diagnoses high-level temporal and social reasoning in short films, our model hits \textbf{44.8\%}, beating the strongest baseline (Video-KTR) by $3.2\%$ and sitting just behind proprietary models like Gemini-2.5-Pro (45.0\%). For knowledge-intensive tasks, we scores \textbf{68.5\%} on \textbf{MMVU (MC)} using noticeably less RL training data, showing our effective design towards complex multimodal reasoning. Crucially, these reasoning gains don't come at the cost of general comprehension: we hit \textbf{62.6\%} on general video benchmarks ahead of the previous best at 60.3\%.
\vspace{-10pt}

\begin{wrapfigure}{r}{0.50\textwidth}
\centering
\captionsetup{type=table}
\caption{Performance comparison with other RL methods. We use the same backbone for fairness, and $\dagger$ denotes our reimplementation results using the same data configurations.}
\vspace{-8pt}
\label{tab:rl_comparison}
\renewcommand{\arraystretch}{1.15}
\setlength{\tabcolsep}{2.5mm}
\resizebox{\linewidth}{!}{%
\begin{tabular}{l|cccc}
    \toprule
    \rowcolor[gray]{.9}
    \textbf{Method} & \textbf{VideoMME} & \textbf{VideoMMMU} & \textbf{MMVU} & \textbf{Video-Holmes} \\ 
    \midrule
    \multicolumn{5}{c}{\textit{Qwen2.5-VL-7B}} \\
    \midrule
    \hspace{2mm} +PAPO$^\dagger$ & 61.0 & 45.7 & 64.3 & 41.5 \\ 
    \hspace{2mm} +Video-OPSD& 58.6 & 49.9 & - & 36.2 \\ 
    \hspace{2mm} \textbf{+Ours} & \textbf{62.6} & \textbf{53.6} & \textbf{68.5} & \textbf{44.8} \\ 
    \midrule
    \multicolumn{5}{c}{\textit{Qwen3-VL-4B}} \\
    \midrule
    \hspace{2mm} +PAPO$^\dagger$ & 60.2 & 54.5 & 64.8 & 41.5 \\ 
    \hspace{2mm} +Video-OPSD & 60.7 & 52.9 & - & 41.8 \\ 
    \hspace{2mm} \textbf{+Ours} & \textbf{63.1} & \textbf{61.0} & \textbf{67.0} & \textbf{42.7} \\ 
    \bottomrule
\end{tabular}%
}
\vspace{-10pt}
\end{wrapfigure}

\paragraph{Performance comparison with existing RL methods.}
As reported in Table~\ref{tab:rl_comparison}, our method consistently outperforms all RL-based approaches, including the adapted PAPO and Video-OPSD, across both backbone settings. The largest gains appear on knowledge-intensive benchmarks: VideoMMMU (+6.5 over PAPO on Qwen3-VL-4B) and MMVU (+4.2 over PAPO on Qwen2.5-VL-7B). These results confirm that answer-relevant token selection and temporal dynamics modeling effectively incorporate useful knowledge while filtering noise.

\subsection{Ablation Study}

\paragraph{Effect of multi-role token selection.} To validate the efficacy of our multi-role token selection strategy, we construct several ablation variants by systematically ablating individual scoring signals. Specifically, we exclude either the answer-aware components ($\Delta H^{c}_{t}$ and $J^{c}_{t}$) or the visual-aware counterparts, while also introducing an entropy-only baseline that allocates credit solely based on decoding uncertainty. As reported in Table~\ref{tab:score_ablation}, the integration of these complementary roles yields consistent performance gains across diverse datasets and evaluation capabilities. These findings indicate that no single metric is sufficient to fully capture the critical factors underlying effective reasoning. Notably, while the entropy-based credit allocation narrows the performance gap with our method on temporally sensitive benchmarks in VideoMME, it exhibits substantial degradation on reasoning-oriented benchmarks. This discrepancy may arise because enhancing visual exploration capability facilitates better cross-frame semantic association, but may also introduce visual noise in reasoning tasks, leading to visual ambiguity and performance deterioration.
\paragraph{Component effects of self-diagnostic evolving modeling.} As shown in Figure~\ref{tab:other_setting}, \textbf{(1)} removing $A_i$ degrades performance across all benchmarks (largest drops: 4.0\% on VideoMME, 3.7\% on Video-Holmes), confirming that it stabilizes policy optimization by filtering low-quality rollouts and preventing suboptimal reasoning reinforcement. 
\begin{wrapfigure}{r}{0.54\textwidth}
\centering
\captionsetup{type=table}
\caption{Ablation study on Qwen2.5-VL-7B.}
\label{tab:other_setting}
\renewcommand{\arraystretch}{1.15}
\setlength{\tabcolsep}{2.5mm}
\resizebox{\linewidth}{!}{%
\begin{tabular}{l|cccc}
    \toprule
    \rowcolor[gray]{.9}
    \textbf{Method} & \textbf{VideoMME} & \textbf{VideoMMMU} & \textbf{MMVU} & \textbf{Video-Holmes} \\ 
    Full Model & 62.6 & 53.6 & 68.5 & 44.8 \\ 
    \midrule
    \multicolumn{5}{c}{\textit{Effect of Advantage Gating}} \\
     \midrule
    w.o. $A_{i}$ Gating & 58.6\textsubscript{$\downarrow$4.0} & 51.0\textsubscript{$\downarrow$2.6} & 64.8\textsubscript{$\downarrow$3.7} & 41.1\textsubscript{$\downarrow$3.7} \\ 
    \midrule
    \multicolumn{5}{c}{\textit{Effect of Curriculum}} \\
    \midrule
    Broad $\rightarrow$ Narrow & 59.1\textsubscript{$\downarrow$3.5} & 49.7\textsubscript{$\downarrow$3.9} & 65.4\textsubscript{$\downarrow$3.1} & 42.3\textsubscript{$\downarrow$2.5} \\ 
    Fixed & 60.6\textsubscript{$\downarrow$2.0} & 51.2\textsubscript{$\downarrow$2.4} & 65.6\textsubscript{$\downarrow$2.9} & 42.1\textsubscript{$\downarrow$2.7} \\ 
    \midrule
    \multicolumn{5}{c}{\textit{Optimization Strategy}} \\
    \midrule
    Re-weighting (Softmax) & 60.5\textsubscript{$\downarrow$2.1} & 50.0\textsubscript{$\downarrow$3.6} & 66.5\textsubscript{$\downarrow$2.0} & 42.1\textsubscript{$\downarrow$2.7} \\ 
    \midrule
    \multicolumn{5}{c}{\textit{Key Frame Selection Strategy}} \\
    \midrule
    $\mathcal{R}_{r}\text{-}focus$ & 61.8\textsubscript{$\downarrow$0.8} & 52.8\textsubscript{$\downarrow$0.8} & 67.6\textsubscript{$\downarrow$0.9} & 44.6\textsubscript{$\downarrow$0.2} \\ 
    $\mathcal{R}_{w}\text{-}focus$ & 61.0\textsubscript{$\downarrow$1.6} & 52.5\textsubscript{$\downarrow$1.1} & 67.2\textsubscript{$\downarrow$1.3} & 41.5\textsubscript{$\downarrow$3.3} \\ 
    \bottomrule
\end{tabular}%
}
\end{wrapfigure} 
\textbf{(2)} Reverse curriculum variants (Narrow $\rightarrow$ Broad) underperform the Broad $\rightarrow$ Narrow setting by 3.5\% on VideoMME, indicating that global-to-local learning misaligns with the model's evolving capacity. Our progressive curriculum, by contrast, balances visual exploration and reasoning reliability by mastering local frame-level reasoning before expanding to longer temporal horizons. 
\textbf{(3)} The re-weighting (softmax) baseline underperforms our contrastive self-distillation by 2.1\% on VideoMME and 3.6\% on VideoMMMU, showing that token weight amplification along training dynamics is more susceptible to visual noise. \textbf{(4)} $\mathcal{R}_{r}$-focus, $\mathcal{R}_{w}$-focus, and their set difference yield comparable results, suggesting that video information attended during incorrect reasoning is not entirely noise and highlighting the inherent redundancy of video.

\subsection{Analysis}

\paragraph{Effect on video hallucination.}
As shown in Table~\ref{tab:videohalluc}, DyCPO consistently improves all hallucination metrics on VidHalluc, validating its effectiveness in suppressing spurious visual noise. It outperforms the w/o $J^{ans}/\Delta H_{ans}$ variant by +1.1 on ACH MCQ (90.6 vs. 89.5), demonstrating that multi-role dependence steers contrastive learning toward answer-relevance mining. Compared to the ``Only Entropy'' ablation, DyCPO achieves a +5.4 gain on STH (93.5 vs. 88.1), reflecting its ability to anchor spatio-temporal grounding and prevent hallucinations of non-existent objects or temporal relationships.

\begin{wraptable}{r}{0.48\textwidth}
\centering
\caption{Performance comparison on video hallucination benchmark \textit{VidHalluc}.}
\vspace{-8pt}
\label{tab:videohalluc}
\resizebox{\linewidth}{!}{%
\begin{tabular}{lcccc}
\toprule
\multicolumn{1}{c}{\multirow{2}{*}{\textbf{Method}}} & \multicolumn{2}{c}{\textbf{ACH}} & \multicolumn{1}{c}{\multirow{2}{*}{\textbf{TSH}}} & \multicolumn{1}{c}{\multirow{2}{*}{\textbf{STH}}} \\
\cmidrule(lr){2-3}
 & \textbf{Binary QA} & \textbf{MCQ} & & \\
\midrule
w/o $J^{ans}/\Delta H_{ans}$    & 90.4 & 89.5 & \textbf{94.8}   & 91.9 \\
Only Entropy & 89.8 & 89.3 & 93.6 & 88.1 \\
Ours    &  \textbf{91.4} & \textbf{90.6} & \textbf{94.8} & \textbf{93.5}  \\
\bottomrule
\end{tabular}%
}
\end{wraptable}

\begin{figure*}[t]
  \centering
  \includegraphics[width=\textwidth]{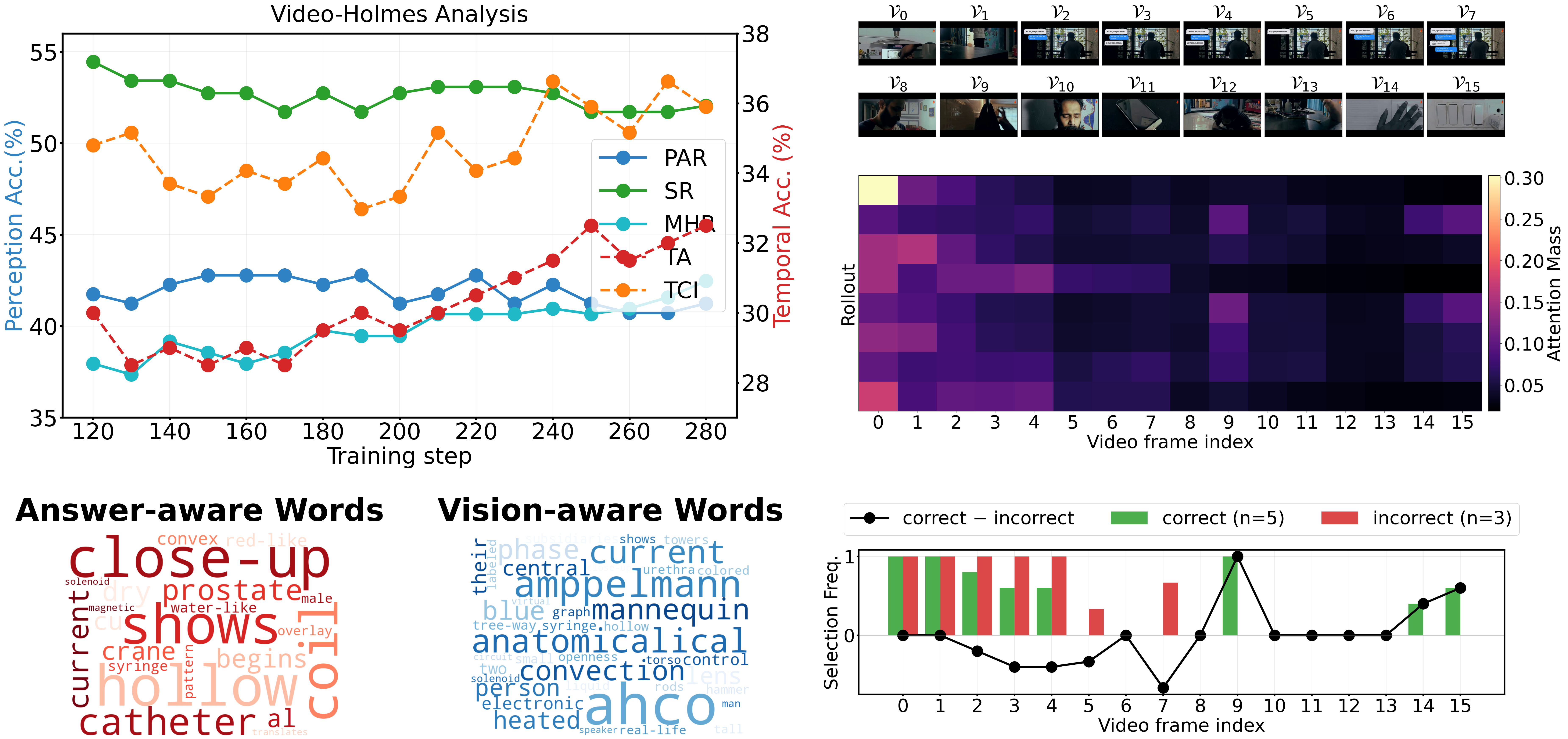}
  \vspace{-8pt}
  \caption{\textbf{Left to right:} (a) Steady rise in temporal metrics (TA, TCI) with stable perception (MHR). (b) Word distribution comparison between answer-aware and vision-aware tokens. (c) Detailed qualitative rollout and fine-grained visual diagnosis on Video-Holmes. The attention maps visualize the frame-level focus of each rollout, where lighter colors indicate higher attention weights. The bar chart at the bottom compares the frequency with which correct and erroneous rollouts attend to specific frames, while the black curve depicts the divergence between them.} 
  \label{fig:combined_three}
\end{figure*}

\paragraph{Effect of curriculum strategy.} As shown in Figure~\ref{fig:combined_three}(a), where TCI stands for \textit{Temporal Causal Inference} and TA denotes \textit{Timeline Analysis}, we observe that the introduction of curriculum learning progressively enhances the model's temporal reasoning capabilities (TA from 28.5\% $\rightarrow$ 31.6\%), while perceptual abilities remain stable. This suggests that our strategy of gradually expanding the visual receptive field from local keyframes enables the model to simultaneously learn local cues and global dependencies.

\paragraph{Visual- and answer-sensitive token distributions.} 
To intuitively illustrate the distinct preferences of visual-aware and answer-aware tokens, we visualize their respective distributions via word clouds in Figure~\ref{fig:combined_three}(b). Tokens with high vision-aware sensitivity concentrate on semantic and color attributes, such as ``blue'', ``heated'', and ``person'', whereas answer-aware tokens are dominated not only by nouns and attribute words but also by conclusive function words like ``aligns'', ``suggests'', and ``followed'', which predominantly reveal conclusions or final answers.

\begin{wrapfigure}{r}{0.35\textwidth} 
    \vspace{-14pt}
    \centering
    \includegraphics[width=\linewidth]{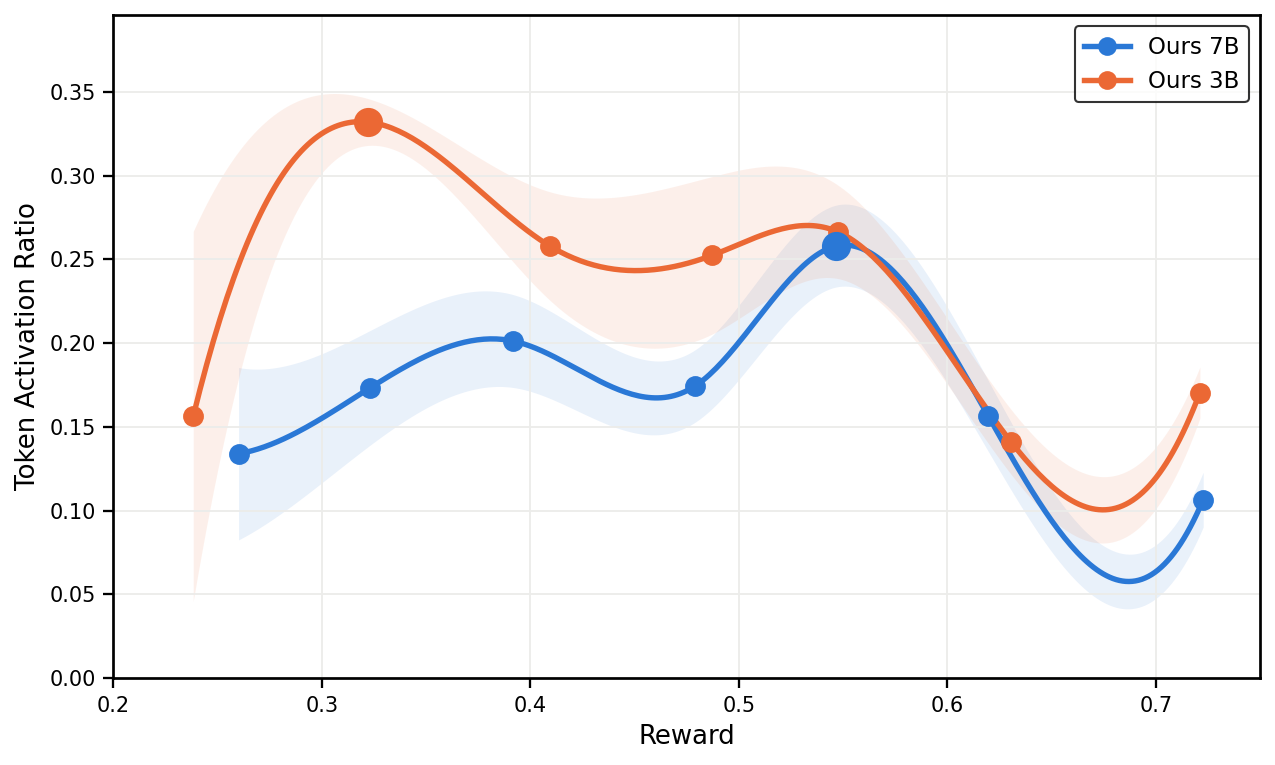}
    \vspace{-18pt}
    \caption{Consistent correlation between reward and token activation fraction across training steps in 7B and 3B models.}
    \label{fig:token_activation_fraction}
\end{wrapfigure}

\paragraph{Visualization for rollout-contrastive visual anchors.} 
To intuitively interpret the behavior of the proposal model, we visualize the keyframes selected via our rollout-contrastive mechanism. As illustrated in Figure~\ref{fig:combined_three}(c), for the question ``What does the disappearance of wall reflection when pressing the switch at 0:13 imply?'', the correct rollout predominantly attends to frames 0--4, while the incorrect rollout notably neglects frames 14, and 15. This indicates that the critical visual cues for resolving the query are the switch-related elements in frames 9, 14, and 15, whereas the early frames merely provide contextual background without decisive evidential value. Our self-diagnostic counterfactual strategy effectively guides the model to attend to these answer-critical visual cues, thereby enhancing reasoning performance. 

\paragraph{Relationship between reward dynamics and token activation ratio.}

We investigate the coupling between reward magnitude and token activation ratio---the fraction of tokens contributing to $\mathcal{L}_{\text{MRC}}$ gradients. Since $\mathcal{L}_{\text{MRC}}$ activates only on rollouts with positive advantages ($A_i > 0$, Eq.~\ref{eq:loss_func}), the mean reward directly reflects the number of informative rollouts per group. As shown in Figure~\ref{fig:token_activation_fraction}, the activation ratio exhibits an M-shaped (bimodal) pattern as reward increases. This non-monotonicity stems from two failure modes at opposite difficulty extremes: (1) overly simple questions yield deterministic ``all-correct'' rollouts, where zero intra-group variance forces all advantages toward zero under GRPO's advantage normalization, eliminating gradients and collapsing token-wise credit assignment. (2) Conversely, prohibitively hard prompts produce near-zero success rates, leaving $\mathcal{L}_{\text{MRC}}$ without valid activation signals from positive-advantage rollouts and similarly suppressing token engagement. Therefore, maintaining rewards within an intermediate band of balanced difficulty is essential to prevent advantage vanishing and signal sparsity, thereby maximizing token activation and the optimization efficacy of $\mathcal{L}_{\text{MRC}}$.

\section{Conclusion}
In this paper, we empirically investigate the distinct roles that individual tokens play during video reasoning and propose a multi-role token selection mechanism coupled with a self-diagnostic rollout-contrastive intervention paradigm. Specifically, we introduce a proposal-policy collaborative learning framework operating at the token level. We first quantify the inherent limitations of high-entropy tokens in video understanding and reveal the functional complementarity between visual-aware and answer-aware activations through a systematic token distribution analysis. We further demonstrate that our self-diagnostic curriculum masking strategy can better and more reliably stimulate the model's ability to capture fine-grained visual concepts and temporal dynamics, yielding robust gains under challenging conditions. Extensive experiments across multiple backbones, including ablations on scoring metrics, masking strategies, and curriculum directionality, as well as thorough comparisons with on-policy distillation methods, consistently demonstrate the effectiveness and superiority of our approach, establishing a new state of the art on key video reasoning benchmarks.

\clearpage

\bibliography{ref}
\bibliographystyle{iclr2027_conference}

\appendix
\addtocontents{toc}{\protect\setcounter{tocdepth}{2}}

\clearpage
\renewcommand{\contentsname}{Appendix}
\section{More Implementation Details}
\paragraph{Models.}

We employ Qwen2.5-VL-7B, Qwen3-VL-2B, and Qwen3-VL-4B as our base
models for detailed ablative analysis. The Qwen2.5-VL models are initialized from a VideoRFT SFT checkpoint, while the Qwen3-VL models start from their official weights.
By default, the base model for benchmark comparison refers to Qwen2.5-VL-7B to ensure fairness.
All variants share the same GRPO configuration and differ only in the perception objective;
we consider the following main model variants and default configurations:
\begin{itemize}
\item \textbf{GRPO}: GRPO~\citep{shao2024deepseekmath} baseline with symmetric
clipping factors $\epsilon_l = \epsilon_h = 0.2$, no reference KL penalty
($\beta$ removed, i.e. \texttt{use\_kl\_loss=False} and \texttt{use\_kl\_in\_reward=False}),
GRPO advantage without per-group standard-deviation normalization, and zero entropy bonus.
\item \textbf{Ours}: Given the
on-policy rollout, the student's own attention over layers $L$ selects $K$
keyframes, on which a keyframe-driven contrastive perception objective $\mathcal{L}_{MRC}$ is computed. The perception score fuses four our defined metrics---vision and answer
Jensen--Shannon divergence together with the absolute entropy shift $|\Delta H|$
(equal $0.25$ weights)---with fusion weight $\alpha$, entropy-gate strength $\eta$,
temperature $\tau=1$, and hard occlusion negatives. \ours introduces \emph{no} EMA
self-distillation teacher; the keyframes only drive the CPL perception term, which is
added to the policy loss with coefficient $\lambda$ under advantage gating.
\end{itemize}

\begin{table}[ht]
\centering
\setlength{\tabcolsep}{1.6mm}
\small
{%
\begin{tabular}{@{}l|cccccccc@{}}
\toprule
\textbf{Model} & \textbf{Frames} & $K_{2}$ & $\bm{\lambda}$ & $\bm{\alpha}$ & $\bm{\eta}$ & \textbf{Layers} $L$ & $\bm{\beta}$ & $\bm{\varepsilon_{l}, \varepsilon_{h}}$ \\
\midrule
\textbf{Qwen2.5-VL GRPO-3B}   & 16 & -      & 0    & -    & -    & -                  & 0.0 & 0.2, 0.2 \\
\textbf{Qwen2.5-VL GRPO-7B}   & 16 & -      & 0    & -    & -    & -                  & 0.0 & 0.2, 0.2 \\
\textbf{Qwen2.5-VL GRPO-7B}   & 24 & -      & 0    & -    & -    & -                  & 0.0 & 0.2, 0.2 \\
\textbf{Qwen2.5-VL GRPO-7B}   & 32 & -      & 0    & -    & -    & -                  & 0.0 & 0.2, 0.2 \\
\textbf{Qwen2.5-VL Ours-3B}  & 16 & 4$\to$8 & 0.05 & 0.5 & 0.01 & Removing last five layers & 0.0 & 0.2, 0.2 \\
\textbf{Qwen2.5-VL Ours-7B}  & 16 & 4$\to$8 & 0.05 & 0.5 & 0.01 & Removing last five layers & 0.0 & 0.2, 0.2 \\
\textbf{Qwen2.5-VL Ours-7B} & 24 & 6$\to$12 & 0.05 & 0.5 & 0.01 & Removing last five layers & - & 0.2, 0.2 \\
\textbf{Qwen2.5-VL Ours-7B} & 32 & 8$\to$16 & 0.05 & 0.5 & 0.01 & Removing last five layers & - & 0.2, 0.2 \\
\midrule
\textbf{Qwen3-VL GRPO-2B}     & 16 & -      & 0    & -    & -    & -                  & 0.0 & 0.2, 0.2 \\
\textbf{Qwen3-VL GRPO-4B}     & 16 & -      & 0    & -    & -    & -                  & 0.0 & 0.2, 0.2 \\
\textbf{Qwen3-VL Ours-2B}    & 16 & 4$\to$8 & 0.05 & 0.5 & 0.01 & wo injection layer        & 0.0 & 0.2, 0.2 \\
\textbf{Qwen3-VL Ours-4B}    & 16 & 4$\to$8 & 0.05 & 0.5 & 0.01 & wo injection layer        & 0.0 & 0.2, 0.2 \\

\bottomrule
\end{tabular}%
}
\caption{\textbf{Hyperparameter configurations} for different models.
$K_{1}$ and $K_{2}$ respectively refers to the ratio of selected response token for keyframes localization and the number of attention-selected keyframes, annealed linearly over training by a
curriculum; $\lambda$ is the MRC perception coefficient (we set to $0$ in GRPO setting),
$\alpha_{1}$ and $\alpha_{2}$ the balacedd weight of scoring fusion weight, $\eta$ the entropy-gate strength, and
$L$ the transformer layers whose response$\to$frame attention is pooled for
keyframe selection. All models remove the reference KL penalty ($\beta$) for better exploration.} 
\label{tab:training_config}
\end{table}

Table~\ref{tab:training_config} summarizes the hyperparameter configurations for the different model variants. For our method, we employ hard occlusion negatives alongside an answer-conditioned hard negative (defined as the rollout with the lowest accuracy within the group) and option-text answers serving as privileged information. The keyframe count $K_{1}$ follows a linear curriculum ($4 \to 8$ for 16-frame inputs and $8 \to 16$ for 32-frame inputs), which scales the temporal receptive field of the perception module according to the input frame budget. Regarding the attention layer set $\mathcal{L}$, we utilize a fixed subset for Qwen2.5-VL, whereas for Qwen3-VL---which employs DeepStack to inject visual features across all layers---we select all layers except those directly receiving visual injections.

\paragraph{Training.}
We conduct RLVR on all model variants using the following typical response format, where
reasoning steps are enclosed in \texttt{\textless think\textgreater\textless/think\textgreater}
and the final answer in \texttt{\textbackslash boxed\{\}}. Each model is trained for
1 epoch on our 3K video-QA training set with a learning rate of $1\mathrm{e}{-6}$, frames
sampled uniformly from each video at $100352$ pixels ($128$ visual
tokens per frame), a maximum prompt length of $4096$ and response length of $512$. During RL policy optimization, we employ a rollout group size of $n=8$ generated responses per prompt, maintaining a batch size of $8$.

\section{Additional Experiments}
\paragraph{Generalization to extended temporal horizons.}
\label{app:long_video_analysis}
Long-video understanding poses severe challenges for multimodal models due to overwhelming frame redundancy and high computational costs. To assess whether our method maintains effectiveness over extended temporal contexts, we evaluate on LVBench~\citep{lvbench}. Table~\ref{tab:lvbench} presents the comparative results. Our method establishes a new state-of-the-art with 40.67\% average accuracy, outperforming GRPO (40.03\%) and entropy-only (39.70\%) baselines. The advantage is particularly pronounced in dynamic content categories: we achieve substantial improvements in Sport (+0.83\% over GRPO), Self-Media (+2.84\%), and Live (+1.15\%). Although GRPO slightly leads in Documentary and TV categories, our method exhibits more balanced performance across the spectrum.

\begin{table}[htbp]
  \centering
  \caption{Performance comparison on different subtasks in LVBench.}
  \label{tab:lvbench}
  \begin{tabular}{lccccccc}
    \toprule
    Model & Avg. & Cartoon & Documentary & Live & Self-Media & Sport & TV \\
    \midrule
    GRPO & 40.03  & 34.28 & \textbf{45.18} & 39.69 & 40.06 & 42.50 & \textbf{40.40} \\
    Entropy-only & 39.70  & 34.63 & 43.15 & 38.93 & 41.01 & 42.08 & 39.60 \\
    \rowcolor{blue!10} Ours & \textbf{40.67} & \textbf{35.34} & 43.15 & \textbf{40.84} & \textbf{42.90} & \textbf{43.33} & 39.20 \\
    \bottomrule
  \end{tabular}
\end{table}

\begin{table*}[ht]
\centering
\small
\setlength{\tabcolsep}{8pt}
\caption{Performance comparison of models trained with different input video frames across five benchmarks.}
\label{tab:frame_wise_comparison}
\resizebox{\linewidth}{!}{%
\begin{tabular}{cccccccc}
\toprule
\textbf{Frames} & \textbf{Method} & \textbf{MMVU} & \textbf{VSI} & \textbf{VideoMME} & \textbf{Video-Holmes} & \textbf{VideoMMMU} & \textbf{Avg.} \\
\midrule

\multirow{2}{*}{$f=24$}
& GRPO    & 66.7 & 36.6 & 61.9 & 43.8 & 50.0 & 51.80 \\
& \textbf{Ours}    & \textbf{68.8}\inc{2.1} & \textbf{39.7}\inc{3.1} & \textbf{63.9}\inc{2.0} & \textbf{45.2}\inc{1.4} & \textbf{51.3}\inc{1.3} & \textbf{53.78}\inc{1.98} \\
\midrule

\multirow{2}{*}{$f=32$}
& GRPO    & 66.2 & 36.7 & 61.5 & 42.4 & 51.3 & 51.62 \\
& \textbf{Ours}    & \textbf{69.0}\inc{2.8} & \textbf{40.2}\inc{3.5} & \textbf{64.8}\inc{3.3} & \textbf{45.2}\inc{2.8} & \textbf{52.8}\inc{1.5} & \textbf{54.40}\inc{2.78} \\

\bottomrule
\end{tabular}%
}
\end{table*}
\paragraph{Training with more frames.}
To investigate the impact of input frame capacity during RL training, we evaluate models trained under different video frames ($f=24, 32$). As shown in Table~\ref{tab:frame_wise_comparison}, we have two key observations:

\textbf{(1) Scaling gains from denser video frame inputs:} Increasing the training frame count from $f=24$ to $f=32$ yields modest but consistent gains. Specifically, our method improves from $63.9\%$ to $64.8\%$ on VideoMME and from $51.3\%$ to $52.8\%$ on VideoMMMU, indicating that exposure to richer temporal contexts during RL training enhances the model's capacity to reason over complex visual semantics.

\textbf{(2) Robustness of designed components:} Under both frame configurations, our method consistently outperforms the standard GRPO baseline across diverse benchmarks. The performance advantage is particularly pronounced on reasoning-oriented benchmarks like \textbf{Video-Holmes} ($42.4\% \rightarrow 45.2\%$) and spatial perception benchmark \textbf{VSI} ($36.7\% \rightarrow 40.2\%$) with $\textit{f = 32}$ video input. This indicates that our multi-role selection effectively activates valuable visual clues inherent in denser frames, ensuring that RL optimization focuses on discriminative, answer-relevant visual tokens. Meanwhile, the self-diagnostic curriculum facilitates the model's ability to search, integrate, and analyze multiple clues before drawing an accurate conclusion.

\begin{table}[ht]
\centering
\caption{\textbf{Feedback source ablation.} Comparative performance on three video benchmarks (Qwen2.5-VL-7B, 3K QA training). The final row indicates our default setting.}
\label{tab:feedback_ablation}
\small
\setlength{\tabcolsep}{3.5pt}
\rowcolors{2}{verylightgray}{white}
\resizebox{\linewidth}{!}{%
\begin{tabular}{llcccc}
\toprule
\textbf{Positive} $r^+$ & \textbf{Negative} $r^-$ 
  & \textbf{MMVU} & \textbf{VSI.} & \textbf{VideoMME} & \textbf{Avg.} \\
\midrule
Peer rollout (right full) & Peer rollout (wrong full)                & 64.9 & 34.0 & 60.3 & 53.1 \\
Peer rollout (right full) & Peer rollout (answer only)        & 65.5 & 34.8 & 60.4 & 53.6 \\
Ground truth answer  & Peer rollout (wrong full)                & 66.2 & 35.8 & 61.4 & 54.5 \\
Ground truth answer  & Peer rollout (random wrong answer) & 67.3 & 37.0 & 61.8 & 55.4 \\
\midrule
Ground truth answer  & Peer rollout (answer only)         & \textbf{68.5} & \textbf{39.2} & \textbf{62.6} & \textbf{56.8} \\
\bottomrule
\end{tabular}%
}
\end{table}

\paragraph{Performance with different answer-aware feedback.}
We investigate the impact of incorporating the answer as privileged information to guide token-wise credit assignment. Table~\ref{tab:feedback_ablation} reveals three main findings: First, utilizing the ground-truth answer as the positive reference ($r^+$) substantially outperforms relying solely on peer full-length rollouts (average 55.6$\%$ vs. 53.4$\%$). This performance gap likely stems from noisy or erroneous tokens present in seemingly correct reasoning chains, which affects the quality of positive anchor.
Second, regarding the negative reference ($r^-$), conditioning on privileged full-length rollout information underperforms directly providing the wrong answer. This indicates that concise error signals can refine the multi-role token set selection and yield more robust contrastive signals for training. Third, sampling negative answers from random incorrect choices rather than actual erroneous rollouts degrades performance. This suggests that effective continual learning requires information derived from the model's intrinsic states and feedback, further validating our design philosophy.

\paragraph{Parametric sensitivity to $\bm \alpha$.}
\begin{wraptable}{r}{0.5\textwidth}
  \centering
  \caption{\textbf{Ablation on hyperparameters $\alpha_{vis}$ and $\alpha_{ans}$}. Performance comparison on VideoMMU and Video-Holmes benchmarks.}
  \label{tab:alpha_ablation}
  \small
  \setlength{\tabcolsep}{4pt}
  \begin{tabular}{ccccc}
    \toprule
    $\alpha_{vis}$ & $\alpha_{ans}$ & \textbf{VideoMMU} & \textbf{Video-Holmes} & \textbf{Avg.} \\
    \midrule
    0.25 & 0.25 & \textbf{61.0} & \textbf{42.7} & \textbf{51.9} \\
    0.35 & 0.15 & 57.9 & 42.2 & 50.1 \\
    0.15 & 0.35 & 58.4 & 42.2 & 50.3 \\
    \bottomrule
  \end{tabular}
\end{wraptable}
\textbf{}
We investigate the impact of the loss weighting vector $\bm{\alpha} = [\alpha_{vis}, \alpha_{ans}]$, which balances the trade-off between the visual-perception and answer-aware components. As shown in Table~\ref{tab:alpha_ablation}, a balanced configuration ($\alpha_{vis} = \alpha_{ans} = 0.25$) consistently achieves superior performance across all benchmarks, whereas any skewed weight distribution leads to noticeable degradation. Specifically, when the answer-aware component heavily dominates visual perception (e.g., shifting the ratio to $0.35 : 0.15$), overall accuracy drops universally, with the most pronounced decline occurring on reasoning-intensive tasks. This behavior stems from the intrinsic bias of answer-dominated tokens: they tend to concentrate on definitive conclusions or logical transition cues, thereby circumventing the diverse visual exploration within intermediate chain-of-thought trajectories.

\paragraph{Sensitivity analysis on hyperparameters $K_1$ and $K_2$.}

\begin{wraptable}{r}{0.7\textwidth}
  \vspace{-12pt}
  \centering
  \caption{(a) Effect of window size ($K_1$) comparing fixed sizes with adaptive selection. (b) Effect of the number of selected tokens for video frame localization ($K_2$).}
  \label{tab:window_ablation}
  \small
  \setlength{\tabcolsep}{3pt}
  
  \begin{minipage}[t]{0.48\linewidth}
    \centering
    \subfloat[$K_1$: Window Size]{
    \begin{tabular}{ccccc}
      \toprule
      \textbf{Fix 2} & \textbf{Fix 4} & \textbf{Fix 6} & \textbf{Fix 8} & \textbf{Adaptive} \\
      \midrule
      59.8 & 61.8 & 62.1 & 61.4 & \textbf{62.6} \\
      \bottomrule
    \end{tabular}
    }
  \end{minipage}%
  \hfill
  \begin{minipage}[t]{0.48\linewidth}
    \centering
    \subfloat[$K_2$: Number of Windows]{
    \begin{tabular}{ccccc}
      \toprule
      \textbf{15} & \textbf{20} & \textbf{25} & \textbf{30} & \textbf{40} \\
      \midrule
      61.9 & \textbf{62.6} & 62.1 & 61.5 & 59.3 \\
      \bottomrule
    \end{tabular}
    }
  \end{minipage}
  \vspace{-8pt}
\end{wraptable}

We evaluate the impact of hyperparameters $K_1$ (window size) and $K_2$ (number of selected windows) on overall performance, as depicted in Table~\ref{tab:window_ablation}. 
As shown in Table~\ref{tab:window_ablation}(a), increasing the window size from 2 to 6 gradually improves performance from 59.8\% to 62.1\%, while the adaptive selection strategy achieves the best result of 62.6\%, demonstrating the effectiveness of dynamically adjusting window sizes. Regarding $K_2$ in Table~\ref{tab:window_ablation}(b), performance follows an inverted U-shape with respect to the number of selected windows, reaching an optimal peak at $K_2=20$ with 62.6\%. A smaller $K_2$ (e.g., 15) risks omitting essential visual context, whereas an excessively large $K_2$ (e.g., 40) introduces redundant frame noise that degrades performance to 59.3\%. These findings confirm that our method achieves optimal performance with adaptive window sizing and moderate window counts, ensuring robustness across different configurations.

\paragraph{Effect of high-attention frames from different layers on training.}
Motivated by the findings in~\citep{DBLP:conf/naacl/ZhangQSYYXWGTY25}, which demonstrate that attention tend to converge on task-relevant semantic regions in earlier-to-intermediate layers but progressively diverge in the deepest layers, we hypothesized that the terminal layers may introduce representational noise detrimental to precise visual grounding. To verify this, we conducted additional ablation studies in which we systematically excluded the terminal layers (specifically, the last five layers) when computing the attention flow. Our empirical results strongly corroborate this hypothesis: pruning the final few layers consistently yields the optimal model performance, whereas relying exclusively on these terminal layers generally results in different degree of degradation.
This provides a stronger premise for localizing key frame videos from rollouts.

\begin{table*}[ht]
\centering
\small
\setlength{\tabcolsep}{14pt}
\caption{Ablation study on selecting high-attention tokens/frames driven by different layer focus across five benchmarks. The base model is Qwen2.5-VL-7B.}
\label{tab:layer_attention_ablation}
\resizebox{\linewidth}{!}{%
\begin{tabular}{cccccc}
\toprule
\textbf{Attention Focus (Layer Group)} & \textbf{MMVU} & \textbf{VSI} & \textbf{VideoMME} & \textbf{Video-Holmes} & \textbf{VideoMMMU} \\
\midrule
Low Layers \textit{(e.g., L1–L8)}       & 66.8 & 36.6 & 61.6 & 43.6 & 51.1 \\
Mid Layers \textit{(e.g., L9–L20)}      & 67.0 & 38.8 & 62.1 & 44.2 & 52.8 \\
Last Few Layers \textit{(e.g., L21–L32)} & 67.2 & 37.8 & 61.9 & 44.8 & 51.1 \\
\midrule
\textbf{Ours (Removing Last Five Layers)}   & \textbf{68.5} & \textbf{39.2} & \textbf{62.6} & \textbf{44.8} & \textbf{53.6} \\
\bottomrule
\end{tabular}%
}

\end{table*}

\paragraph{Perturbation strategies.}
Unlike image-based contrastive perception methods such as PAPO, we adopt diverse perturbations within videos, which require dynamic temporal understanding. Given an on-policy rollout $\mathbf{o} = (o_1, \dots, o_M)$, the student's own response-to-frame attention selects a keyframe set $\mathcal{K} \subseteq \{1, \dots, F\}$, where $F$ denotes the total number of input frames. All negative-view perturbations act exclusively on the target frames $\mathcal{T}$ (by default $\mathcal{T} = \mathcal{K}$, optionally extended to all $F$ frames), ensuring that corruption aligns with the temporal support actually utilized by the policy. The positive view $\mathcal{V}^{+}$ applies mild photometric jittering (brightness, contrast, saturation), while the negative view $\mathcal{V}^{-}$ removes visual information from $\mathcal{T}$. We consider three perturbation variants, all applied in pixel space with patch size $s \times s$ ($s = 14$).

\paragraph{(1) Random masking.}
We design two stochastic occlusion strategies that preserve temporal order while degrading appearance:
\begin{itemize}
    \item \textbf{Random region occlusion}: Scatters $n$ random rectangles per target frame $t \in \mathcal{T}$. Each rectangle's side lengths are sampled independently as
    \begin{align}
    f_h, f_w \sim \mathcal{U}(\rho_{\min}, \rho_{\max}), \quad
    r_h = \lfloor f_h H \rfloor, \quad r_w = \lfloor f_w W \rfloor,
    \end{align}
    where $\mathcal{U}$ denotes the uniform distribution. Rectangles are placed at uniformly random locations and filled with per-pixel RGB noise sampled from $\mathcal{U}(\min_t, \max_t)$ (the frame's dynamic range). We use $n = 50$, $\rho_{\min} = 0.05$, and $\rho_{\max} = 0.15$.

    \item \textbf{Single-block noise} (\texttt{region\_noise}): Masks one contiguous rectangle covering a fraction $\rho = 0.5$ of each frame, filled with uniform noise.
\end{itemize}

\paragraph{(2) Attention-aware masking.}
These strategies implement ``semantic-aware'' corruptions targeting regions where the model focuses its attention. To identify critical regions, we compute an attention map $\mathbf{a}_t \in \mathbb{R}^{H_b \times W_b}$ by summing the cross-attention weights $a^{(l)}_{m,v}$ (defined in Eq.~\ref{eq:dep-low-entropy}) over visual tokens in frame $t$, where $H_b = H/s$ and $W_b = W/s$ after merging. We select the top-$\phi$ tokens with highest attention scores, denoted $\mathcal{S}_t = \operatorname{Top}\text{-}\lfloor \phi H_b W_b \rfloor(\mathbf{a}_t)$, and define their bounding box $\mathcal{B}_t$ as the \emph{attention hotspot}.

We propose two hotspot-based corruption strategies:
\begin{itemize}
    \item \textbf{Attention-focused occlusion}: Places all $n$ noise rectangles strictly inside the hotspot $\mathcal{B}_t$ ($\phi = 0.2$ in our experiments), attacking regions the model relies on most heavily.

    \item \textbf{Curriculum occlusion}: Gradually expands the corruption scope. Each rectangle is placed randomly in the full frame with probability $p$ (where $p$ increases from $0$ to $1$ during training), and inside the hotspot $\mathcal{B}_t$ otherwise:
    \begin{align}
    \operatorname{region}(t) =
    \begin{cases}
    \text{full frame} & \text{with probability } p, \\
    \mathcal{B}_t & \text{with probability } 1-p.
    \end{cases}
    \end{align}
    This anneals from purely attention-focused corruption ($p \approx 0$, hard but localized) to purely random corruption ($p = 1$, easier but global), enabling progressive adaptation to diverse noise patterns.
\end{itemize}

\paragraph{(3) Temporal order destruction.}
To disrupt sequential dependencies, we shuffle the keyframe slots using a derangement permutation $\pi$, where $\mathcal{V}^{-}_t = \mathcal{V}_{\pi(t)}$ for $t \in \mathcal{K}$. This breaks the temporal order while keeping the spatial content of each frame intact, which is different from permuting tokens within a single frame.

In our training, we apply temporal shuffling and masking-based corruptions independently with a probability of $0.5$. Table~\ref{tab:masking_strategy} shows the results of different perturbation strategies. We find that \textbf{random region occlusion} works best overall. This aligns with previous studies on masked visual modeling, which show that random masking is often better than attention-based masking. We explain this with two reasons:
\begin{itemize}[leftmargin=*, nosep]
\item \textbf{Attention-focused masking removes a single salient block.} This leaves too much context in the negative view, allowing the model to guess the answer from the surroundings. As a result, the contrastive signal is weaker on appearance-heavy tasks like MMVU ($67.04$ vs.\ $68.16$) and VideoMME ($61.91$ vs.\ $62.37$). It only helps slightly on VSI ($38.55$ vs.\ $37.67$), where forcing the model to look away from the most attended region is beneficial.
\item \textbf{Scattering many small random rectangles degrades information more evenly.} Instead of removing an entire object, this approach creates a harder and better-calibrated negative sample. The curriculum variant starts with attention-focused masking and inherits its weakness, so it does not outperform plain random occlusion.
\end{itemize}

\begin{table}[ht]
\centering\small
\begin{tabular}{@{}l|ccc@{}}
\toprule
\textbf{Negative-view perturbation strategy} & \textbf{MMVU} & \textbf{VSI} & \textbf{VideoMME} \\
\midrule
Random region occlusion          & \textbf{68.16} & \textbf{38.55} & \textbf{62.37} \\
Attention-focused occlusion & 67.04 & 37.67 & 61.91 \\
Curriculum (attention$\to$full region) & 66.88 & 37.55 & 62.03 \\
\bottomrule
\end{tabular}
\caption{\textbf{Impact of the negative-view (masking) strategy} on our method. All rows share
the same base model, data, and hyperparameters and differ only in how the negative view is corrupted. }
\label{tab:masking_strategy}
\end{table}

\section{Training dynamics}

\paragraph{Reward, response length, and loss analysis.}
We analyze the optimization behavior of our method in Figure~\ref{fig:training_dynamics}, highlighting three key observations regarding the interplay between perception supervision and policy learning.

\textbf{(1) The perception objective exhibits stable convergence with an initial adaptation phase.}
As shown in Figure~\ref{fig:training_dynamics}(a), the MRC loss experiences a transient spike within the first $\sim\!50$ steps as the model adapts to the contrastive views. Following this brief adjustment, the loss decreases monotonically and stabilizes near $0.03$. This trajectory confirms that the auxiliary perception objective converges smoothly without causing destructive interference with the primary policy-gradient updates.

\textbf{(2) The contrastive objective effectively enforces representation separation.}
Figure~\ref{fig:training_dynamics}(b) decomposes the embedding space dynamics by tracking distances from the anchor to positive ($d_{pos}$) and negative ($d_{neg}$) views. Throughout training, $d_{pos}$ remains consistently and significantly lower than $d_{neg}$ (maintaining a gap of $\sim\!0.015$). This clear separation demonstrates that the model reliably pulls answer-consistent views closer while pushing perturbed negatives away, validating the efficacy of the contrastive term in structuring the visual-textual representation space.

\textbf{(3) Perception supervision is reward-aligned yet promotes concise reasoning.}
Crucially, adding the perception objective does not compromise reinforcement learning performance. In Figure~\ref{fig:training_dynamics}(c), our method's reward trajectory closely aligns with the GRPO baseline, indicating that the auxiliary loss is effectively reward-neutral. However, the response-length dynamics in Figure~\ref{fig:training_dynamics}(d) reveal a distinct advantage: while GRPO struggles to compress outputs (stabilizing around $140\text{--}160$ tokens), our method drives a more aggressive reduction in length, converging to a more concise $\sim\!120$ tokens. We attribute this to the perception objective encouraging the model to discard redundant reasoning steps while retaining essential evidence-grounded content, leading to more efficient inference without sacrificing accuracy.

\begin{figure*}[t]
  \centering
  \includegraphics[width=0.97\linewidth]{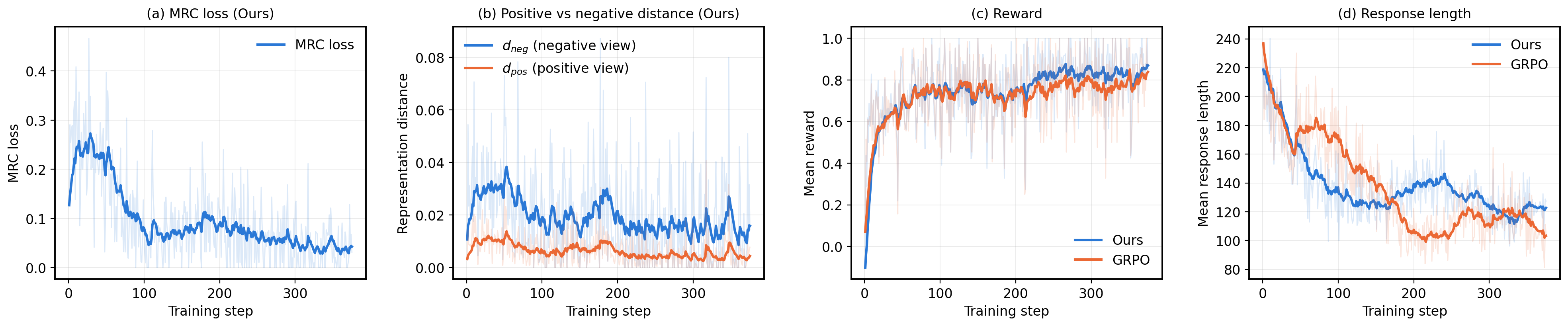}
\vspace{-5pt}
  \caption{\textbf{Left to right}. (a) Convergence curve of $\mathcal{L}_{MRC}$ loss. (b) Distance metrics of positive vs. negative samples relative to the anchor. (c) Reward trajectory. (d) Variations in reasoning chain length.}
  \label{fig:loss_reward_length}
\end{figure*}
\label{app:attention_analysis}

\paragraph{Training efficiency and computational overhead.}
To evaluate the practical overhead introduced by our method, we report a step-wise time breakdown across various model architectures ($2\text{B}\sim7\text{B}$) in Table~\ref{tab:training_efficiency}. Compared to the standard GRPO baseline, our method incurs a modest increase in total step time, ranging from $+22.3\%$ ($+8.5\text{s}$ on Qwen3-VL-2B) to $+39.9\%$ ($+18.3\text{s}$ on Qwen3-VL-4B). Importantly, rollout generation times remain virtually unchanged across all scales, indicating that our method introduces no latency bottleneck during context sampling. Crucially, a closer inspection reveals that the additional step latency is almost entirely concentrated in the actor update phase, which aligns closely with the forward inference time of the teacher model (e.g., $18.1\text{s}$ update increment vs. $17.1\text{s}$ teacher forward time on 7B). This confirms that our mechanism maintains modest computational efficiency without imposing unmanageable optimization burdens.
\begin{table*}[t]
\centering
\small
\caption{\textbf{Per-step training efficiency and computational overhead breakdown across different model scales.} All time metrics are measured in seconds ($s$) per training step. The extra computational overhead introduced by our method is modest and is almost entirely dominated by the forward pass of the teacher model during policy updates.}
\label{tab:training_efficiency}
\setlength{\tabcolsep}{9pt}
\renewcommand{\arraystretch}{1.15}
\resizebox{\linewidth}{!}{%
\begin{tabular}{llcccccc}
\toprule
\multirow{2}{*}{\textbf{Base Model}} & \multirow{2}{*}{\textbf{Method}} & \multicolumn{4}{c}{\textbf{Per-Step Time Breakdown (s)}} & \multicolumn{2}{c}{\textbf{Overhead vs. GRPO}} \\
\cmidrule(lr){3-6} \cmidrule(lr){7-8}
 & & \textbf{Rollout Gen.} & \textbf{Actor Update} & \textbf{Extra Fwd.} & \textbf{Avg. / Step} & \textbf{Absolute ($\Delta t$)} & \textbf{Relative ($\Delta$)} \\
\midrule
\multirow{2}{*}{Qwen3-VL-2B}   & GRPO & 16.0 & 14.9 & 0.0 & 38.1 & -- & -- \\
                                       & Ours& 16.5 & 22.0 & 7.5 & 46.6 & +8.5s & +22.3\% \\
\midrule
\multirow{2}{*}{Qwen3-VL-4B}   & GRPO & 15.6 & 20.5 & 0.0 & 45.9 & -- & -- \\
                                       & Ours & 18.2 & 34.3 & 11.9 & 64.2 & +18.3s & +39.9\% \\
\midrule
\multirow{2}{*}{Qwen2.5-VL-7B} & GRPO & 19.8 & 30.2 & 0.0 & 62.6 & -- & -- \\
                                       & Ours & 20.1 & 48.3 & 17.1 & 81.6 & +19.0s & +30.4\% \\
\bottomrule
\end{tabular}%
}
\end{table*}

\section{Additional analysis on related work}
We present a fine-grained comparison with Video-KTR to underscore our distinct contributions. Our method differs from Video-KTR in three pivotal aspects:
\noindent\textbf{(1) Token selection vs. reward reweighting.} Unlike Video-KTR, which directly integrates score-based reweighting into the GRPO objective, we eschew this strategy. Our empirical analysis reveals that such reweighting yields suboptimal performance on specific datasets, as it tends to amplify visual noise and bias the model towards the base policy. Instead, we leverage scores exclusively to identify salient tokens without explicitly modulating their relative importance, thereby enhancing robustness against visual noise.
\noindent\textbf{(2) Multi-dimensional token analysis.} We perform a distributional quantification of diverse token types with positional preferences, introducing answer-aware perturbation as a critical metric. Furthermore, we trace the origins of high entropy to low-entropy visual activations and characterize the decoding patterns associated with these tokens. Additionally, our self-diagnostic curriculum masking mechanism stabilizes the training process, delivering consistent performance gains across multiple benchmarks.
\noindent\textbf{(3) Superior performance.} Our method demonstrates superior efficacy, consistently outperforming Video-KTR across a comprehensive range of evaluation metrics.

\end{document}